%% file: preprint.tex
\documentclass{article} 
\usepackage{iclr2027_conference,times}

\input{math_commands.tex}

\usepackage{hyperref}
\usepackage{url}
\usepackage{graphicx}

\usepackage{adjustbox}
\usepackage{siunitx}

\newcommand{\falconscore}[2]{%
  \mbox{%
    \makebox[2.8em][r]{$#1$}\,$\pm$\,%
    \makebox[2.2em][l]{$#2$}%
  }%
}

\hypersetup{
    colorlinks=true,
    allcolors=black,
    pdfborder={0 0 0}
}

\usepackage{xcolor}
\usepackage{float}
\usepackage{booktabs}

\usepackage{titletoc}

\newcommand{\name}{\footnotesize\textsf{NeuroLens}}
\newcommand{\figname}{\scriptsize\textsf{NeuroLens}}

\title{NeuroLens: Learning Latent Embeddings of Neural Semantics from Chronic Recordings}

\author{}

\usepackage{titlesec}
\titlespacing*{\section}{0pt}{*1.3}{*1.3}
\titlespacing*{\subsection}{0pt}{*1.2}{*1.2}
\titlespacing*{\paragraph}{0pt}{0.5ex}{1.em}
\iclrfinalcopy 

\begin{document}
\pagestyle{plain}
\thispagestyle{plain}

{\fontsize{17}{20}\selectfont\scshape
NeuroLens: Learning Latent Embeddings of Neural Semantics from Chronic Recordings
\par}

\vspace{1.5em}

{\fontsize{10}{11}\selectfont
\noindent
\textbf{Hanrui Lyu}$^{1,*,\dagger}$ \quad
\textbf{Baiyuan Chen}$^{2,*}$ \quad
\textbf{Tianshu Tan}$^{3,*}$ \\
\textbf{Matthew R. Whiteway}$^{4}$ \quad
\textbf{Maxwell D. Melin}$^{5}$ \quad
\textbf{Ji Xia}$^{4}$ \quad
\textbf{Linyang He}$^{4}$ \\
\textbf{Bradly C. Stadie}$^{1}$ \quad
\textbf{Anne Churchland}$^{5}$ \quad
\textbf{Liam Paninski}$^{4}$ \quad
\textbf{Yizi Zhang}$^{6,\dagger}$

\vspace{0.8em}

\noindent
$^{1}$Northwestern University \quad
$^{2}$University of Cambridge \quad
$^{3}$Johns Hopkins University \\
$^{4}$Columbia University \quad
$^{5}$University of California, Los Angeles \quad
$^{6}$Stanford University

\vspace{0.4em}

\noindent
$^{*}$Equal contribution
}

\makeatletter
\def\blfootnote{\xdef\@thefnmark{}\@footnotetext}
\makeatother

\blfootnote{
$^{\dagger}$Corresponding authors:
\href{mailto:hanruilyu2029@u.northwestern.edu}
{hanruilyu2029@u.northwestern.edu},
\href{mailto:yizzhang@stanford.edu}
{yizzhang@stanford.edu}
}

\begin{abstract}
Understanding how neural activity represents higher-order cognition and how these representations evolve during learning is a central pursuit in neuroscience. {However, tracking meaningful representational changes remains challenging because recorded neural populations change across days.} Here, we introduce {\name} (\textbf{L}atent \textbf{E}mbeddings of \textbf{N}eural \textbf{S}emantics), a self-supervised model based on the Joint-Embedding Predictive Architecture (LeJEPA). An adaptive encoder maps changing neural populations into a common latent space, where a temporal predictor forecasts future latent states. Across mouse and human intracortical recordings, {\name} learns representations that improve decoding of abstract variables, including decisions and sentence meaning. {In chronic mouse neural recordings during visual decision-making, these representations capture gradual transitions between learning stages and the progressive orthogonalization of neural states associated with different choices. They also reveal distinct timescales of cortical reorganization across tasks during learning. With multi-day pretraining, these representations also support stable decoding across days, including leading performance on the FALCON benchmark. Together, these findings support {\name} as a framework for tracking learning-related representational changes while maintaining stable decoding over long timescales.}
\end{abstract}

\section{Introduction}
Learning continually reshapes the relationship between neural activity and behavior over extended timescales \citep{peters2014emergence}. Understanding this process requires tracking how task-relevant population dynamics emerge, persist, and reorganize across days \citep{wang2026dynamics}. Advances in chronic recordings increasingly enable such longitudinal measurements \citep{melin2024large,bimbard2025adaptable,skromne2026months}, but electrode drift and changes in the recorded neurons can obscure learning-related changes in neural activity \citep{karpowicz2024few}. {This challenge motivates methods that preserve learning-related representational changes while accommodating changes in the recorded population.}

Existing approaches address recording nonstationarity through statistical and deep-learning-based neural manifold alignment \citep{karpowicz2025stabilizing, gallego2020long}, recalibration-based test-time adaptation \citep{fan2023plug,feghhi2026time}, and transformer-based in-context learning \citep{le2026spint}. {These methods improve decoding stability across recording days, but stable decoding alone does not establish whether learned representations preserve meaningful changes during learning. This motivates a general-purpose framework that accommodates changing neural populations while retaining evolving task-relevant structure.}

Self-supervised learning (SSL) provides a foundation for learning such representations beyond a specific decoding task. However, many existing approaches reconstruct observed neural activity, which may retain recording-specific details alongside behaviorally relevant structure \citep{ye2021representation, zhang2024towards}. For example, probe drift can alter which neurons are recorded even when the underlying population dynamics remain stable. Reconstruction-based models must account for these changes in the observations, although they may be unrelated to the neural computation of interest \citep{rule2020stable}. Joint-embedding predictive architectures (JEPAs) instead predict masked or future representations in latent space \citep{assran2023self}. This objective emphasizes predictable structure without requiring the model to reproduce every detail of the recorded activity. {Applied to chronic recordings, this raises the question of whether latent space prediction can learn neural representations that capture task-relevant changes during learning.}

Building on this principle, we introduce {\name} (\textbf{L}atent \textbf{E}mbeddings of \textbf{N}eural \textbf{S}emantics), a self-supervised model that learns denoised, semantically informative neural latents through causal prediction in latent space \citep{maes2026leworldmodel}. The model integrates two complementary components. An adaptive neural encoder uses cross-attention to accommodate varying neuron counts across days and infers neuron identities from a few calibration trials, mapping each day's neural population into a common latent space. A causal transformer predictor then predicts future latent embeddings, with parameters shared across days to encourage {temporally consistent representations}. These components enable adaptation to changing neural populations while capturing temporal regularities across days.

Across chronic intracortical data in mice \citep{melin2026behavioral} and humans \citep{willett2023high, card2024accurate}, {we show that {\name} learns task-relevant latent geometry that supports improved decoding of choice and sentence meaning. Multi-day pretraining yields temporally consistent representations with stable effective rank, alongside better decoding on future sessions than state-of-the-art (SOTA) baselines. The model also supports  gradient-free few-shot adaptation to unseen neural populations. On the FALCON benchmark's M2 track, {\name} achieves the first place among non-ensemble methods across all three cross-day adaptation settings: zero-shot, few-shot unsupervised, and few-shot supervised, and more than doubles the strongest competing zero-shot baseline's held-out decoding performance. Further analyses show that its neural latents capture the progressive orthogonalization of neural states as animals learn a decision-making task, as well as cortical reorganization during this process. Together, these findings support the use of {\name} to study learning-related population dynamics over time.} 

\section{Related Work}

\paragraph{Large-scale models for neural analysis.} Recent advances have expanded from modeling single-session neural dynamics to large-scale pretraining across sessions, animals, and tasks \citep{andre2026brainwidebench}. Supervised approaches such as POYO \citep{azabou2023unified} and POSSM \citep{ryoo2026generalizable} enable scalable multi-session decoding through spike tokenization combined with PerceiverIO \citep{jaegle2021perceiver} or recurrent state-space models. Self-supervised approaches \citep{ye2021representation,zhang2024towards,zhang2025neural} instead learn representations by masking neural activity across time, neurons, regions and modalities. {These approaches differ in their objectives: supervised models optimize predefined decoding targets, whereas many self-supervised models predict or reconstruct observed neural activity. Beyond task transfer and decoding performance, an important question is whether learned representations remain consistent over time while preserving meaningful changes during learning.}

\paragraph{Self-supervised representation learning for neural activity.} Self-supervised learning enables transferable representations from unlabeled data. Observation-space methods such as GPT \citep{radford2018improving} predict future inputs, while masked autoencoders (MAEs) \citep{he2022masked} reconstruct missing inputs. Although these methods scale effectively, they require modeling low-level details that may be irrelevant to downstream tasks. Joint-Embedding Predictive Architectures (JEPAs) instead predict masked targets in latent space, encouraging higher-level structure \citep{bardes2024revisiting}. I-JEPA \citep{assran2023self} demonstrated semantic visual representations without hand-crafted augmentations, while LeJEPA \citep{balestriero2025lejepa} introduced a theoretically grounded regularizer to prevent representation collapse. {Recently, JEPA-based approaches have extended latent-space prediction to non-invasive neural recordings \citep{guetschel2024s, dong2024brain}, demonstrating its potential for learning neural representations beyond observation space reconstruction. However, these methods, developed for non-invasive data, do not address the distinct challenges of long-term modeling in chronic intracortical recordings.}

\paragraph{Nonstationarity in chronic neural recordings.} Methods for addressing nonstationarity in chronic neural recordings have largely been developed for task-specific decoding and fall into three categories. First, manifold-alignment methods map activity across days into a shared latent space using statistical or deep learning techniques; NoMAD maps activity to a stable reference manifold \citep{karpowicz2025stabilizing}, while CEBRA learns consistent embeddings from temporal context \citep{schneider2023learnable}. Second, test-time recalibration methods continually update decoders, with CORP and DietCORP using language-model-corrected pseudo-labels for online adaptation \citep{fan2023plug}. Third, transformer-based methods, including SPINT and POSSM, use causal prediction objectives \citep{le2026spint,ryoo2026generalizable}. {These approaches offer complementary strategies for maintaining decoding performance across days, including alignment, recalibration, and supervised adaptation. Our focus is on self-supervised representations that support stable decoding across changing neural populations while retaining meaningful representational changes during learning. }

\section{Datasets}
\paragraph{IBL chronic Neuropixels data.} This dataset tracks a mouse learning a visual decision-making task in which it turns a wheel to report whether a stimulus appears on the left or right side of a screen, receiving a reward for correct choices \citep{melin2026behavioral}. Chronic Neuropixels recordings capture spiking activity from hundreds of neurons over 3 months as the mouse progresses from novice to proficient performance. We analyze 53 daily sessions ($\sim$47 hours) from this mouse, including both trial and non-trial (spontaneous) periods. Each session contains 366 trials on average, each spanning 2 seconds around stimulus onset; non-trial periods are divided into contiguous 2-second windows. Spikes are binned at approximately 16.7 ms (60 bins per second), and units with mean firing rates below 1 Hz are excluded. Each recording contains 159$\sim$498 units across approximately 6$\sim$20 anatomical regions. Neural activity is paired with synchronized left- and right-camera videos of the face and upper body. Behavioral variables include trial-level binary choice and time-bin-level wheel speed, whisker motion energy from each camera, and paw and nose speeds from the left camera. 

\paragraph{Human chronic Utah array data.} These datasets contain neural recordings from two participants (T12 and T15) with ALS-induced anarthria as they attempted to speak displayed sentences. For T12, chronic recordings were obtained from four intracortical Utah arrays implanted in the ventral motor cortex \citep{willett2023high}; we retain the 128 channels in area 6v. We analyze 9,680 trials across 23 daily sessions over four months, totaling 16.2 hours. For T15, we analyze 9,498 trials across 45 sessions over 20 months, using all 256 electrodes in the precentral gyrus \citep{card2024accurate}. For both datasets, neural features include threshold-crossing spike counts and spiking-band power (SBP), binned at 20 ms and z-scored to mitigate recording nonstationarity (Appendix~\ref{app:data_preprocessing}). Each sentence is labeled with an unaligned phoneme sequence using a 41-class vocabulary that includes the Connectionist Temporal Classification (CTC) blank token. For semantic decoding, we obtain a 384-dimensional sentence embedding from all-MiniLM-L6-v2 \citep{reimers2019sentence}, such that sentences with similar meanings have similar embeddings.

\section{Methods}
\vspace{-1mm}
{\name} is a self-supervised model that learns task-relevant neural representations by predicting future latent embeddings. Building on the LeJEPA-based temporal prediction framework of LeWorldModel~\citep{maes2026leworldmodel}, it addresses nonstationarity in chronic recordings via a day-adaptive encoder, which maps daily activity into a common latent space by conditioning on neural population size and identity, and a shared autoregressive (AR) predictor, which extracts temporally predictable structure to form a stable subspace across days. Training combines prediction mean squared error (MSE) with sketched isotropic Gaussian regularization (SIGReg) to prevent representation collapse~\citep{balestriero2025lejepa}. At inference, the encoder supports few-shot, gradient-free adaptation to unseen populations. We detail the architecture and training objectives below.

\begin{figure}[!tbp]
    \centering
    \includegraphics[width=\textwidth]{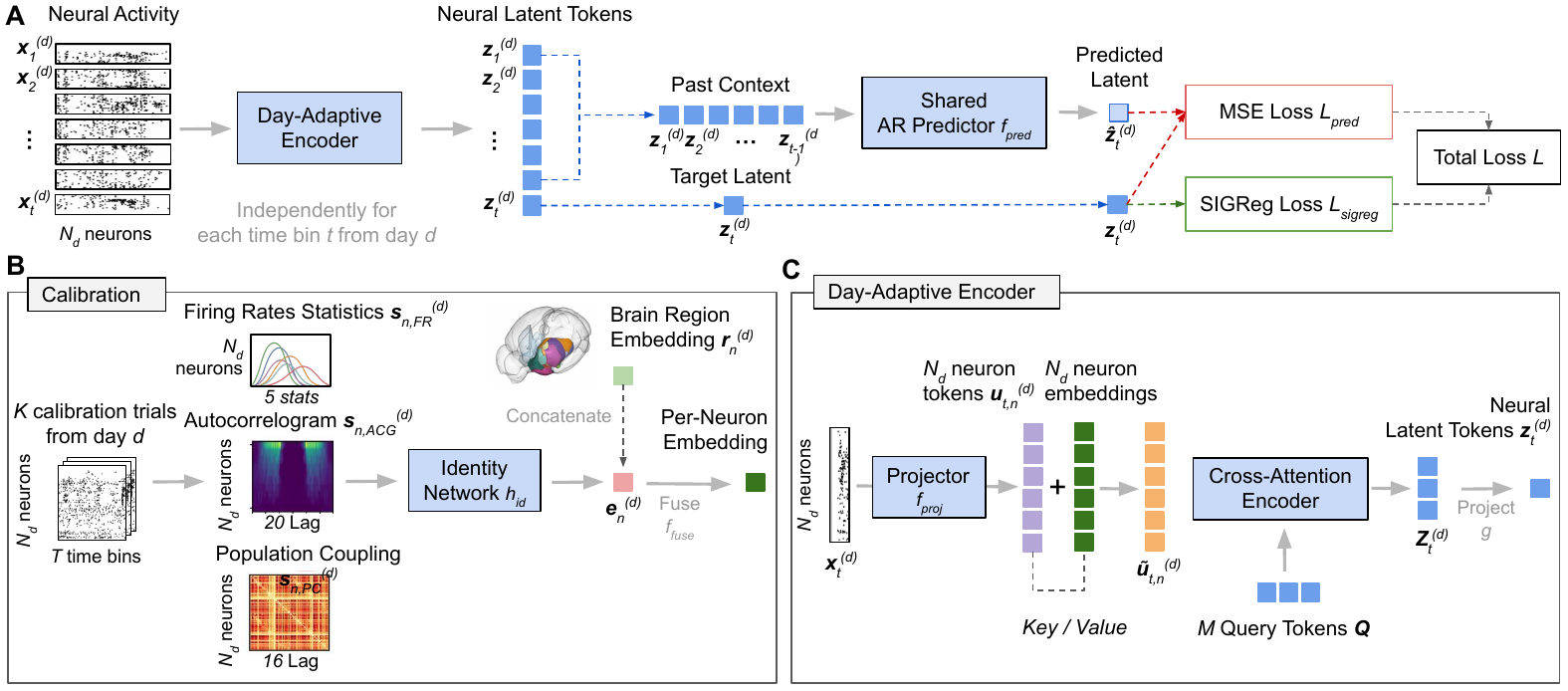}
    
    \caption{\footnotesize \textbf{{\figname} framework.} (A) {\em Model overview.} A day-adaptive encoder maps spike counts at each time bin to latent representations, and an autoregressive (AR) predictor shared across days predicts future latents from past latents. The model is trained using a latent prediction loss together with SIGReg loss to prevent representation collapse. (B-C) {\em Day-adaptive cross-attention encoder based on calibration data.} At each time bin, an MLP independently projects each neuron’s spike count into a neuron token. These tokens serve as the keys and values of the cross-attention module. Because attention is permutation-invariant, we distinguish neuron identities using calibration trials sampled from each recording day. From these trials, we compute firing-rate statistics, autocorrelograms (ACGs), and population coupling to generate a signature for each neuron. A second MLP maps these signatures to neuron embeddings, which are concatenated with brain-region embeddings and added to the corresponding neuron tokens. Learned queries then aggregate the variable-length sequence of neuron tokens through cross-attention, producing a fixed number of latent embeddings for the AR predictor. Processing each time bin independently prevents temporal leakage, while calibration-based neuron embeddings enable adaptation to previously unseen neural populations across recording days.} \label{fig:model}
\end{figure}

\subsection{Model Architecture}\label{sec:model_architecture}

\paragraph{Day-adaptive cross-attention encoder.} To accommodate changes in recorded neurons and their responses across days, we map variable-size neural populations into a common latent space. Let $\mathbf{x}^{(d)}_t \in \mathbb{R}^{N_d}$ denote spike counts for $N_d$ neurons on day $d$ at time bin $t$. A two-layer MLP $f_{\mathrm{proj}}$ with GELU activation maps each spike count to a neuron token $
\mathbf{u}^{(d)}_{t,n}=f_{\mathrm{proj}}\left(x^{(d)}_{t,n}\right)
$, yielding
$
\mathbf{U}^{(d)}_t=\{\mathbf{u}^{(d)}_{t,n}\}_{n=1}^{N_d}
\in\mathbb{R}^{N_d\times H}.
$ We augment each spike token with neuron and region embeddings to obtain $\widetilde{\mathbf{U}}^{(d)}_t$ (defined below). Then, $M$ learned queries $\mathbf{Q}\in\mathbb{R}^{M\times H}$ aggregate neuron tokens via cross-attention, and a linear projection $g$ maps each latent embedding to a scalar:
\begin{equation}
    \mathbf{Z}^{(d)}_t
    =
    \operatorname{CrossAttn}
    \!\left(
        \mathbf{Q},
        \widetilde{\mathbf{U}}^{(d)}_t,
        \widetilde{\mathbf{U}}^{(d)}_t
    \right)
    \in\mathbb{R}^{M\times H}, \qquad
    \mathbf{z}^{(d)}_t = g(\mathbf{Z}^{(d)}_t)
\in \mathbb{R}^{M}.
\label{eq:x_attn}
\end{equation}
Each time bin is encoded independently: $\mathbf{z}^{(d)}_t$ depends only on $\mathbf{x}^{(d)}_t$. This prevents temporal leakage during encoding and leaves temporal modeling to the subsequent AR predictor.

\paragraph{Calibration-based neuron identification.} Cross-attention is invariant to neuron-token order (Eq.~\ref{eq:x_attn}), so spike-count tokens alone do not explicitly encode neuron identity. We therefore characterize each neuron $n$ on day $d$ using a small set of daily calibration trials:
$
    \mathbf{s}^{(d)}_n
=
\operatorname{concat}\!\left(
    \mathbf{s}^{(d)}_{n,\mathrm{FR}},\,
    \mathbf{s}^{(d)}_{n,\mathrm{ACG}},\,
    \mathbf{s}^{(d)}_{n,\mathrm{PC}}
\right), 
$
where FR, ACG, and PC summarize firing-rate statistics, autocorrelograms, and population coupling in 5, 20, and 16 dimensions, respectively (Appendix~\ref{app:neuron_signatures}). A two-layer network $h_{\mathrm{id}}$ with GELU and output LayerNorm maps each signature to a neuron embedding $\mathbf{e}^{(d)}_n$, which is concatenated with a learnable region embedding $\mathbf{r}^{(d)}_n$, projected into token space, and added to the spike token before cross-attention (Eq.~\ref{eq:x_attn}):
\begin{equation}
    \mathbf{e}^{(d)}_n
    =
    h_{\mathrm{id}}\!\left(\mathbf{s}^{(d)}_n\right)
    \in\mathbb{R}^{H'}, \quad
    \widetilde{\mathbf{u}}^{(d)}_{t,n} =
    \operatorname{LayerNorm}\!\left(
        \mathbf{u}^{(d)}_{t,n}
        + f_{\mathrm{fuse}}\!\left(
        \operatorname{concat}\!\left(\mathbf{e}^{(d)}_n, \mathbf{r}^{(d)}_n \right)
        \right)
    \right)
    \in\mathbb{R}^{H}.
    \label{eq:neuron_signature}
\end{equation}
The identity network uses physiological and population signatures instead of fixed neuron-ID embeddings \citep{azabou2023unified,ryoo2026generalizable}, enabling transfer to unseen neurons using few calibration trials. Appendix~\ref{app:ablation_neuron_signatures} shows that such signature-based identities perform comparably to learned ID embeddings.

For human Utah array recordings, however, channels capture mixtures of nearby neurons rather than spike-sorted units. Applying neuron signatures (Eq.~\ref{eq:neuron_signature}) can therefore obscure channel identity and induce representation collapse. We instead forgo the cross-attention encoder in Eq.~\ref{eq:x_attn}: channel features are mapped to the latent space by an MLP encoder, and a hypernetwork conditioned on calibration statistics generates day-specific FiLM~\citep{perez2018film} parameters for the encoder layers to adapt to channel drift (Appendix~\ref{app:channel_drift_stats}). {We summarize the differences between calibration-based neuron identification and day-specific FiLM in Appendix~\ref{app:stats_vs_film} to clarify the rationale behind each method’s design.}

\paragraph{Autoregressive predictor.} We employ a causal transformer $P_\theta$, shared across days, to model the temporal dynamics of encoder latents $\mathbf{z}^{(d)}_{1:T}$ (Eq.~\ref{eq:x_attn}). After adding learned temporal position embeddings, we pass $\mathbf{z}^{(d)}_{1:T}$ through transformer blocks, each applying pre-normalized, causally masked multi-head self-attention followed by a feed-forward network. Each prediction uses only preceding latents:
$
    \widehat{\mathbf{z}}^{(d)}_{t}
    =
    f_{\mathrm{pred}}\!\left(
        P_\theta(\mathbf{z}^{(d)}_{1:t-1})
    \right),
$
where $f_{\mathrm{pred}}$ is a two-layer MLP with intermediate GELU activation and batch normalization. Appendix~\ref{app:model_details} provides architectural details and hyperparameters.

\subsection{Training Objectives}
The training objective combines latent prediction loss and regularization loss:
$
\mathcal{L}
=
\mathcal{L}_{\mathrm{pred}}
+
\lambda\mathcal{L}_{\mathrm{sigreg}},
$
where $\lambda$ controls regularization strength.

\paragraph{Prediction loss.} We minimize the mean-squared error between predicted and target encoder latents to encourage the predictor to capture temporally predictable structure in the latent space:
\begin{equation}
    \mathcal{L}_{\mathrm{pred}}
    =
    \frac{1}{(T-1)M}
    \sum_{t=2}^{T}
    \left\|
        \widehat{\mathbf{z}}^{(d)}_{t}
        -
        \mathbf{z}^{(d)}_{t}
    \right\|_2^2.\label{eq:ar_predictor}
\end{equation}

\paragraph{Regularization loss.} Jointly training the encoder and predictor can yield trivial constant representations. To prevent representation collapse, we use the sketch isotropic Gaussian regularizer (SIGReg) \citep{balestriero2025lejepa}, which matches the characteristic functions of randomly projected encoder embeddings to that of an isotropic Gaussian distribution, encouraging nonzero variance across latent dimensions. For encoder embeddings $\mathbf{z}_{t,b}^{(d)} \in \mathbb{R}^{M}$, where $b$ indexes the $B$ samples in a batch, and random unit vectors $\mathbf{a}_j \in \mathbb{R}^{M}$, define the empirical characteristic function of each projection as 
$\widehat{\phi}_{t,j}(\xi)
    =
    \frac{1}{B}
    \sum_{b=1}^{B}
    \exp\!\left(
        i \xi \,\mathbf{a}_j^\top\mathbf{z}_{t,b}^{(d)}
    \right)$. Using $J$ projections, the regularizer is
\begin{equation}
    \mathcal{L}_{\mathrm{sigreg}}
    =
    \frac{B}{TJ}
    \sum_{t=1}^{T}
    \sum_{j=1}^{J}
    \sum_{k=1}^{K}
    w_k
    \left\|
        \widehat{\phi}_{t,j}(\xi_k)
        -
        e^{-\xi_k^2/2}
    \right\|_2^2,
\end{equation}
where the integral is approximated using Gaussian-weighted trapezoidal quadrature with $K=17$ uniformly spaced points $\xi_k \in [0,3]$ and corresponding weights $w_k$. 

\subsection{Generalization to unseen data after pretraining}\label{sec:adaptation}

We pretrain {\name} on recordings from multiple days, using calibration trials from each day to condition and update the encoder. We then evaluate three strategies for adapting a pretrained model to unseen neural populations. These strategies vary in their calibration data requirements, computational cost, and predictive performance.

\paragraph{Gradient-free (\textit{GF}) few-shot adaptation.} We compute neuron signatures $\mathbf{s}^{(d)}_n$ from calibration trials sampled from an unseen day and apply the pretrained identity network $h_{\mathrm{id}}$ (Eq.~\ref{eq:neuron_signature}) to infer identities for previously unseen neurons and obtain neural embeddings (Fig.~\ref{fig:model}B--C). Cross-attention accommodates varying neuron counts, mapping the new population into the learned latent space without updating model parameters. For neural decoding, we train a linear decoder on the frozen predictor representations $\widehat{\mathbf{z}}^{(d)}_{t}$.

\paragraph{Gradient-based (\textit{GB}) few-shot adaptation.} 
We adapt to data from a new day using sampled calibration trials, updating only the read-in layers of the encoder, including the spike projector $f_{\text{proj}}$, the identity network $h_{\text{id}}$, and the fusion projector $f_{\text{fuse}}$ (Figure~\ref{fig:model}). For neural decoding, we additionally update an attached decoder head. All other parameters, including the encoder cross-attention layers, region embeddings, and the AR predictor, are frozen to preserve the representations and dynamics learned during pretraining.

\paragraph{Full finetuning.} With sufficient data from a new session, we fine-tune the entire model, including the decoder head for neural decoding, when few-shot adaptation may not perform well for tasks such as speech decoding \citep{willett2023high}.

\section{Results}

\subsection{Task-relevant latent geometry supports abstract decoding}\label{sec:high_vs_low_decoding}

Brain–computer interfaces decode movement and speech, but recovering higher-level information such as decisions and sentence meaning remains less explored. We test whether predicting in latent space produces representations that better separate this information and thereby support decoding. In mouse and human recordings, we first examine choice and semantic separation, then assess whether these representational properties translate into improved decoding while preserving movement and phoneme information.

\begin{figure}[!htbp]
    \centering
    \includegraphics[width=\textwidth,height=0.24\textheight]{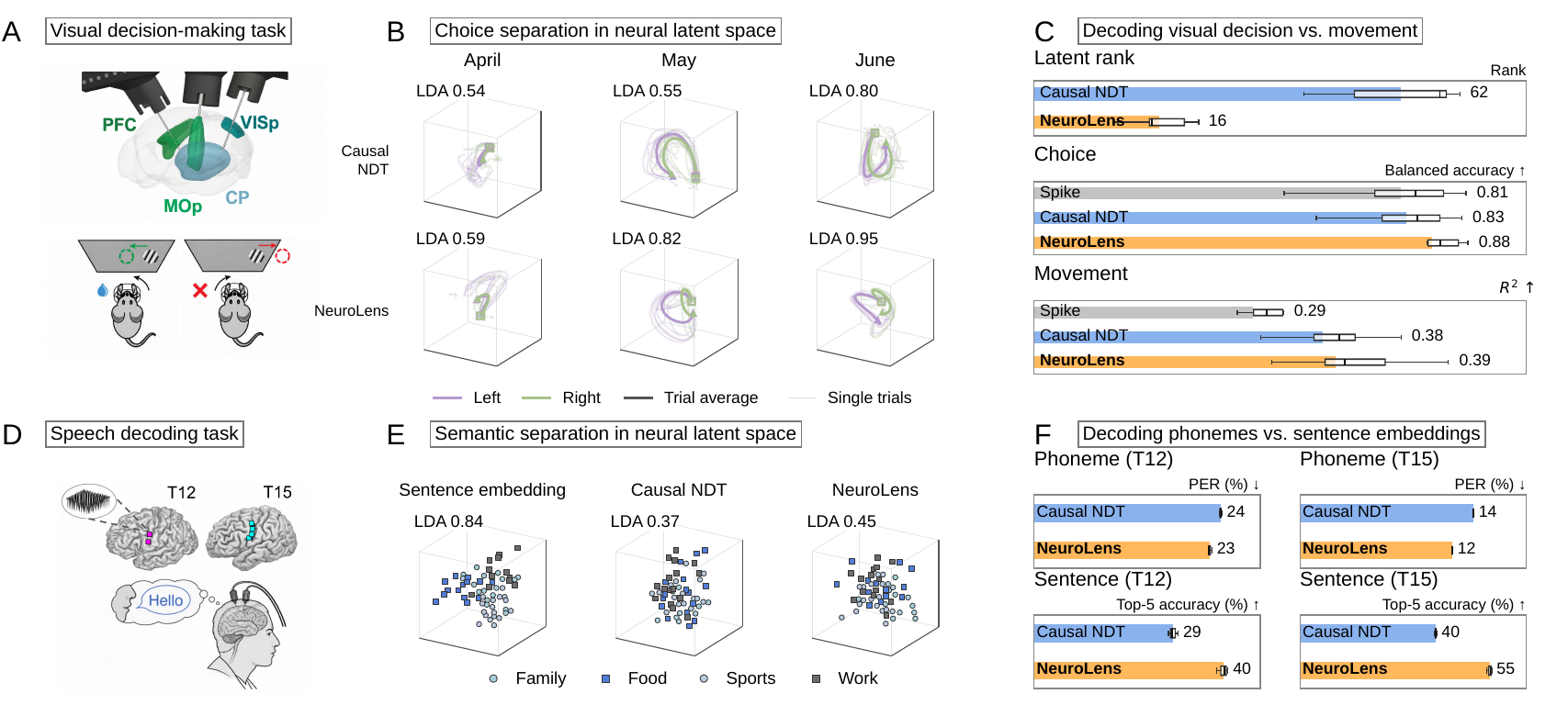}
    
    \caption{\footnotesize \textbf{{\figname} enhances choice and semantic separation in neural latents, alongside improved decoding in mouse and human intracortical recordings.} (A) Multi-area Neuropixels recordings from a mouse reporting stimulus location by turning a wheel. (B) Choice-related latent trajectories separate during learning, with greater separation in {\figname} than in causal NDT. (C) Across 14 test sessions, {\figname} yields lower latent effective rank and better choice decoding, with comparable movement decoding. Both models outperform raw neural activity (\textit{Spike}). (D) Multi-area Utah array recordings from two human participants attempting to speak displayed sentences. (E) Trial-averaged latents show clearer semantic-topic separation in {\figname} than in causal NDT, alongside ground-truth sentence embeddings \citep{reimers2019sentence}. In (B) and (E), latents are projected onto the first three PCs, and separation is quantified by LDA balanced accuracy. (F) {\figname} improves top-5 sentence retrieval accuracy across held-out days while maintaining comparable phoneme decoding. Boxplots summarize five random seeds. Each day’s candidate pool contains 288$\sim$448 sentences, keeping chance level low (1$\sim$2\%).} \label{fig:figure2}
\end{figure}

\paragraph{Baselines and evaluation.} We compare {\name} with causal NDT, which predicts neural activity rather than latents \citep{ye2021representation}, matching latent dimensionality, model capacity, and training setup (Appendix~\ref{app:training_details}). For mouse recordings, we also include a linear decoder trained on raw spikes (\textit{Spike}). We evaluate higher-level decoding of choice and sentence meaning alongside lower-level decoding of movement and phonemes. 

For IBL, we train a separate model for each of 14 sessions and evaluate frozen representations using linear decoders. We measure choice decoding with balanced accuracy and movement decoding with $R^2$. For human recordings, we pretrain a shared model on pooled data from T12 and T15, then fine-tune separate participant-specific models for each decoding task. We evaluate sentence-embedding decoding with top-5 retrieval accuracy and phoneme decoding with phoneme error rate (PER). Data splits, checkpoint selection, and decoder details are provided in Appendix~\ref{app:data} and \ref{app:training_details}.

\paragraph{Choice separation supports decision decoding.} As the mouse learns the task, left- and right-choice trajectories become increasingly separated, with greater separation in {\name} than in causal NDT (Figure~\ref{fig:figure2}B). This separation is quantified by LDA balanced accuracy. Consistent with this representational advantage, {\name} achieves better choice decoding across 14 sessions despite a lower effective latent rank, while maintaining comparable movement decoding (Figure~\ref{fig:figure2}C). Both models outperform raw-spike decoding. Together, these results suggest that {\name} concentrates decision-related information into a more compact representation without sacrificing movement information.

\paragraph{Semantic separation supports sentence decoding.} Human recordings show a parallel pattern. Trial-averaged latents from {\name} separate more clearly by semantic topic than those from causal NDT (Figure~\ref{fig:figure2}E). This improved semantic organization is accompanied by better sentence-embedding decoding, with comparable phoneme decoding (Figure~\ref{fig:figure2}F). Across both datasets, clearer separation of higher-level information is therefore associated with better decoding, supporting the benefit of latent space prediction for learning task-relevant neural representations.

\subsection{Temporally Consistent Latent Geometry Supports Stable Decoding}\label{sec:across_time}

Recording drift challenges stable neural decoding over time. We investigate whether {\name} learns temporally consistent representations with stable dimensionality, and whether these properties support adaptation and decoding on future days.

\begin{figure}[!htbp]
    \centering
    \includegraphics[width=\textwidth]{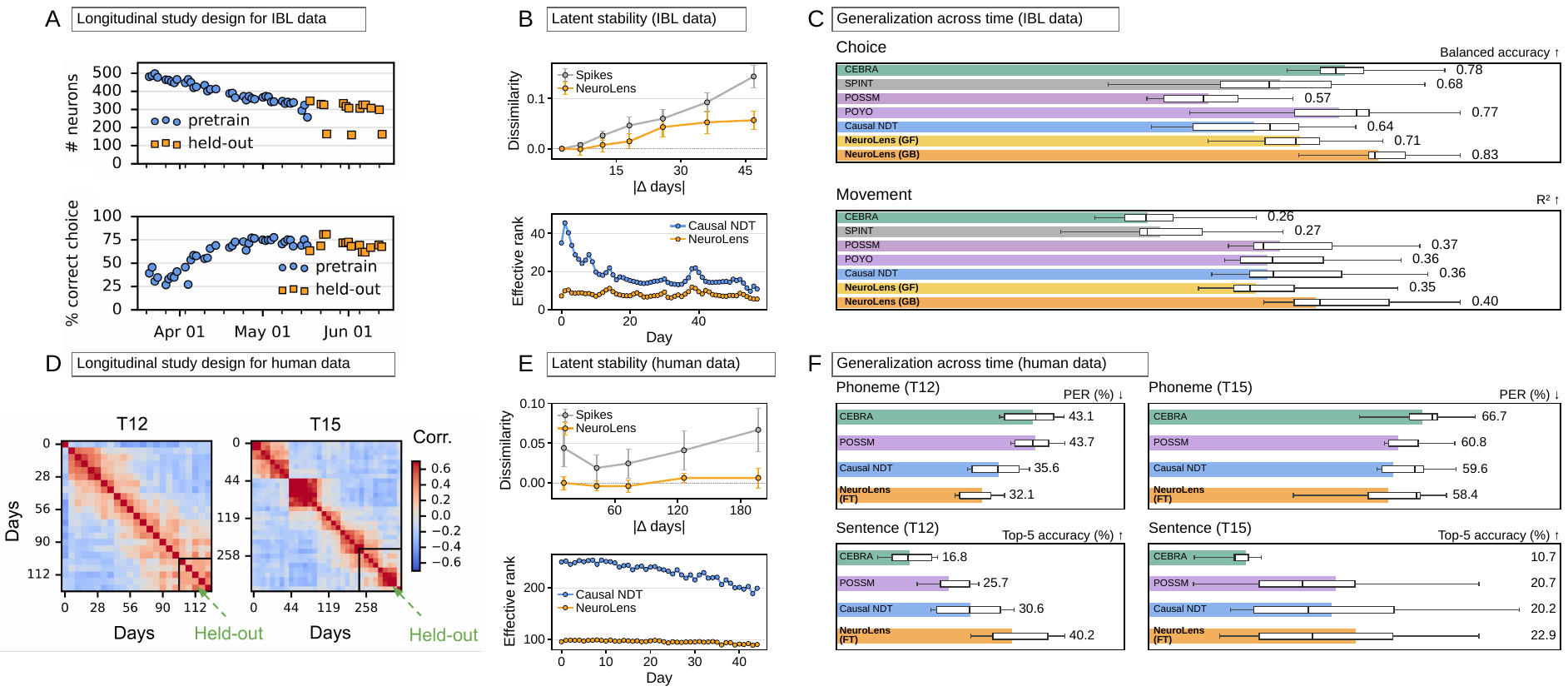}
    \vspace{-4mm}
    \caption{\footnotesize \textbf{{\figname} learns temporally consistent latents with stable effective rank, supporting generalization across time and improved decoding over SOTA baselines.} (A) From April to June, the mouse makes increasingly accurate decisions while the number of simultaneously recorded neurons declines. Models are pretrained on the first 39 days and evaluated on the final 14 days. (B) Across increasing temporal gaps, {\figname} preserves greater representational similarity (Appendix~\ref{app:rsa}) than raw-spike PCA in both mouse (IBL) and human (T15) recordings. (C) Across 14 held-out days, {\figname} outperforms baselines in choice and movement decoding, demonstrating better generalization across time. (D) Human recordings exhibit day-to-day distribution shifts; the final 5 (T12) and 9 (T15) days are held out for evaluation. (E) {\figname} maintains a compact, stable effective rank over time in IBL and T15, whereas causal NDT’s rank declines. (F) With full fine-tuning, {\figname} outperforms baselines in phoneme and sentence-embedding decoding across held-out days. Lower phoneme error rate (PER) and higher top-5 sentence retrieval accuracy indicate better decoding performance.} \label{fig:figure3}
\end{figure}

\paragraph{Baselines and evaluation.} For IBL, we compare {\name} with causal NDT, CEBRA, SPINT, POYO, and POSSM, spanning self-supervised representation learning and supervised decoding \citep{ye2021representation,schneider2023learnable,le2026spint,azabou2023unified,ryoo2026generalizable}. For human recordings, we exclude SPINT because it requires fixed-length windows and POYO because of poor sentence-embedding decoding. Model configurations and training details are provided in Appendix~\ref{app:model_details} and \ref{app:training_details}.

We pretrain IBL models on 39 recording days and evaluate on the following 14 days (Figure~\ref{fig:figure3}A), using separate adaptation, validation, and test sets. We compare gradient-free adaptation using 64 unlabeled calibration trials to infer neuron identities, followed by linear decoder training, with gradient-based adaptation that updates model-specific input layers and decoders while freezing the backbone (Sec.~\ref{sec:adaptation}). For human recordings, we hold out the final 5 days of T12 and 9 days of T15, pretrain on earlier recordings, and fully fine-tune participant- and task-specific models before evaluation (Figure~\ref{fig:figure3}D).

\paragraph{Stable latent structure across days.} We first examine whether {\name} preserves representational structure despite changes in chronic recordings. Representational similarity analysis (RSA; Appendix~\ref{app:rsa}) shows that raw-spike representations become increasingly dissimilar as the gap between recording days grows, whereas {\name} latents remain more consistent in both IBL and T15 (Figure~\ref{fig:figure3}B). Moreover, as the number of recorded neurons declines in IBL, causal NDT's effective latent rank decreases, while {\name} maintains a compact, stable rank across days in both datasets (Figure~\ref{fig:figure3}E). Together, these properties suggest that {\name} provides a consistent representational basis for decoding despite changes in the recorded population.

\paragraph{Improved decoding on future days.} Consistent with this stable latent structure, {\name} achieves the best choice and movement decoding across held-out IBL days (Figure~\ref{fig:figure3}C). Even with gradient-free adaptation, it outperforms SPINT, POSSM, and causal NDT on choice decoding, and CEBRA and SPINT on movement decoding, demonstrating that its pretrained representations remain useful with limited calibration. In human recordings, {\name} also outperforms baselines in phoneme and sentence-embedding decoding on future days after full fine-tuning (Figure~\ref{fig:figure3}F). These findings support temporal consistency and stable latent dimensionality as properties that help sustain decoding across recording days.

\begingroup

\paragraph{FALCON benchmark evaluation setup.}
We further evaluate {\name} on the M2 track of the FALCON competition \citep{karpowicz2024falcon}, a public benchmark that tests generalization from earlier recording days to later, held-out days with limited calibration data. The task is to predict 2-D finger velocity from Utah array spiking activity recorded from a monkey across 13 sessions over 8 days. The first 4 days (7 sessions) are held-in, and the last 4 days (6 sessions) are held-out. For held-in sessions, the first 60\% of neural and behavioral data is released for training; for held-out sessions, only the first 10\% is available for calibration. We evaluate three cross-day adaptation settings: (1) \textit{zero-shot (ZS)}, which uses no held-out calibration data; (2) \textit{few-shot unsupervised (FSU)}, which uses only calibration neural activity to adapt to neural distribution on a new day; and (3) \textit{few-shot supervised (FSS)}, which uses both calibration neural activity and behavioral labels. {\name} flexibly supports all three adaptation settings using a unified model. See Appendix~\ref{app:falcon_M2} for data splits, training, and evaluation details.

\begin{table}[!htbp]
\centering
\footnotesize
\caption{ \footnotesize \textbf{{\figname} markedly outperforms competing baselines in zero-shot and few-shot unsupervised settings for stable decoding on the FALCON M2 benchmark.} Decoding performance is measured by held-out $R^2$ (mean $\pm$ std over the 4 held-out days). Latency is wall-clock decoding time divided by neural recording duration (lower is better). NeuroLens’s normalized latency ranges are below the threshold of 1, supporting real-time use by keeping decoding faster than data acquisition. All methods use 10\% of each held-out session for calibration, except the provided oracle references, which use private labels from 60\% of each held-out session. TCN + GRU ensemble is the only ensemble baseline; all other baselines are non-ensemble models. GF means gradient-free updates; all other methods use gradients computed from the calibration data to update the model. NeuroLens uses GF updates in the FSU setting. $\dagger$ means no associated publication.}
\label{tab:falcon_main}
\setlength{\tabcolsep}{4pt}
\renewcommand{\arraystretch}{0.9}

\begin{adjustbox}{max width=\columnwidth}
\begin{tabular}{@{}l c c c c@{}}
\toprule
\textbf{Method} &
\textbf{\textit{Zero-Shot (ZS)}} &
\textbf{\textit{Few-Shot Unsupervised (FSU)}} &
\textbf{\textit{Few-Shot Supervised (FSS)}} &
\textbf{Latency} \\
\cmidrule(lr){2-2}
\cmidrule(lr){3-3}
\cmidrule(lr){4-4}
\cmidrule(l){5-5}

& \multicolumn{4}{c@{}}{\textbf{\textit{Non-ensemble methods (10\% calibration)}}} \\
\midrule

RNN$^\dagger$
& \falconscore{-0.07}{0.23}
& --
& --
& 0.01 \\

Wiener filter$^\dagger$
& \falconscore{0.06}{0.04}
& --
& --
& 0.08 \\

NoMAD + WF \citep{karpowicz2025stabilizing}
& --
& \falconscore{0.20}{0.10}
& --
& 0.91 \\

CycleGAN + WF \citep{ma2023using}
& --
& \falconscore{0.22}{0.06}
& --
& 0.09 \\

SPINT (GF) \citep{le2026spint}
& --
& \falconscore{0.26}{0.13}
& --
& 0.13 \\

PRISM$^\dagger$
& --
& \falconscore{0.27}{0.13}
& --
& 0.03 \\

OMEN$^\dagger$
& --
& --
& \falconscore{0.41}{0.08}
& 0.04 \\

S5 SSM$^\dagger$
& --
& --
& \falconscore{0.41}{0.11}
& 0.02 \\

NDT2 Multi \citep{ye2023neural}
& --
& --
& \falconscore{0.43}{0.08}
& 0.10 \\

NDT3 \citep{ye2026generalist}
& --
& --
& \falconscore{0.48}{0.06}
& 0.07 \\

TaskFuse$^\dagger$
& --
& --
& \falconscore{0.50}{0.11}
& 0.03 \\

PAR$^\dagger$
& --
& --
& \falconscore{0.53}{0.10}
& 0.32 \\

\addlinespace[1pt]
\textbf{NeuroLens (ours)}
& \falconscore{\mathbf{0.16}}{\mathbf{0.16}}
& \falconscore{\mathbf{0.44}}{\mathbf{0.10}}
& \falconscore{\mathbf{0.53}}{\mathbf{0.08}}
& 0.26 $\sim$ 0.30 \\

\midrule
& \multicolumn{4}{c@{}}{\textbf{\textit{Ensemble methods (10\% calibration)}}} \\
\addlinespace[1pt]

TCN + GRU ensemble$^\dagger$
& --
& --
& \falconscore{0.60}{0.07}
& 0.38 \\

\midrule
& \multicolumn{4}{c@{}}{\textbf{\textit{Private labels oracle references (60\% calibration)}}} \\
\addlinespace[1pt]

NDT2 Multi \citep{ye2023neural}
& --
& --
& \falconscore{0.58}{0.04}
& 0.10 \\

\bottomrule
\end{tabular}
\end{adjustbox}
\end{table}

\paragraph{Leading performance on FALCON with improved zero-shot and few-shot decoding.} {\name} ranks first among non-ensemble methods across all three settings (Table~\ref{tab:falcon_main}). Notably, {\name} achieves remarkable zero-shot performance ($R^2=0.16$), substantially exceeding the strongest ZS baseline ($R^2=0.06$) and approaching NoMAD's FSU performance ($R^2=0.2$) without using any held-out calibration data. With unlabeled calibration data, {\name} improves decoding $R^2$ to 0.44 through gradient-free adaptation, substantially outperforming the strongest FSU baseline ($R^2=0.27$) and surpassing NDT2 Multi's supervised performance ($R^2=0.43$). With labeled calibration data, {\name} achieves an $R^2$ of 0.53, approaching the oracle reference ($R^2=0.58$) while using only 10\% of each held-out session, compared with the oracle's 60\% private labeled calibration data, showing greater data efficiency. These results highlight NeuroLens’s ability to generalize zero-shot and adapt effectively with limited, unlabeled data. 

\paragraph{Decoding latency.} To assess real-time decoding feasibility, we use FALCON's normalized
latency, defined as the total wall-clock time spent generating predictions
divided by the total duration of the evaluated neural recordings. A value
below 1 indicates that decoding is faster than data acquisition on average, and can support real-time decoding. As shown in Table~\ref{tab:falcon_main}, {\name} achieves normalized
latencies of 0.26--0.30 across ZS, FSU, and FSS, meeting this criterion.

\endgroup

\begingroup

\subsection{Latent Geometry Captures Learning-Related Representational Changes}

Although temporal consistency supports stable decoding across days, it may also suppress meaningful changes over longer timescales. We therefore examined whether {\name} captures evolving task-related representations during learning in chronic Neuropixels recordings from one IBL mouse \citep{melin2026behavioral}.

\begin{figure}[!t]
    \centering
    \includegraphics[width=\textwidth]{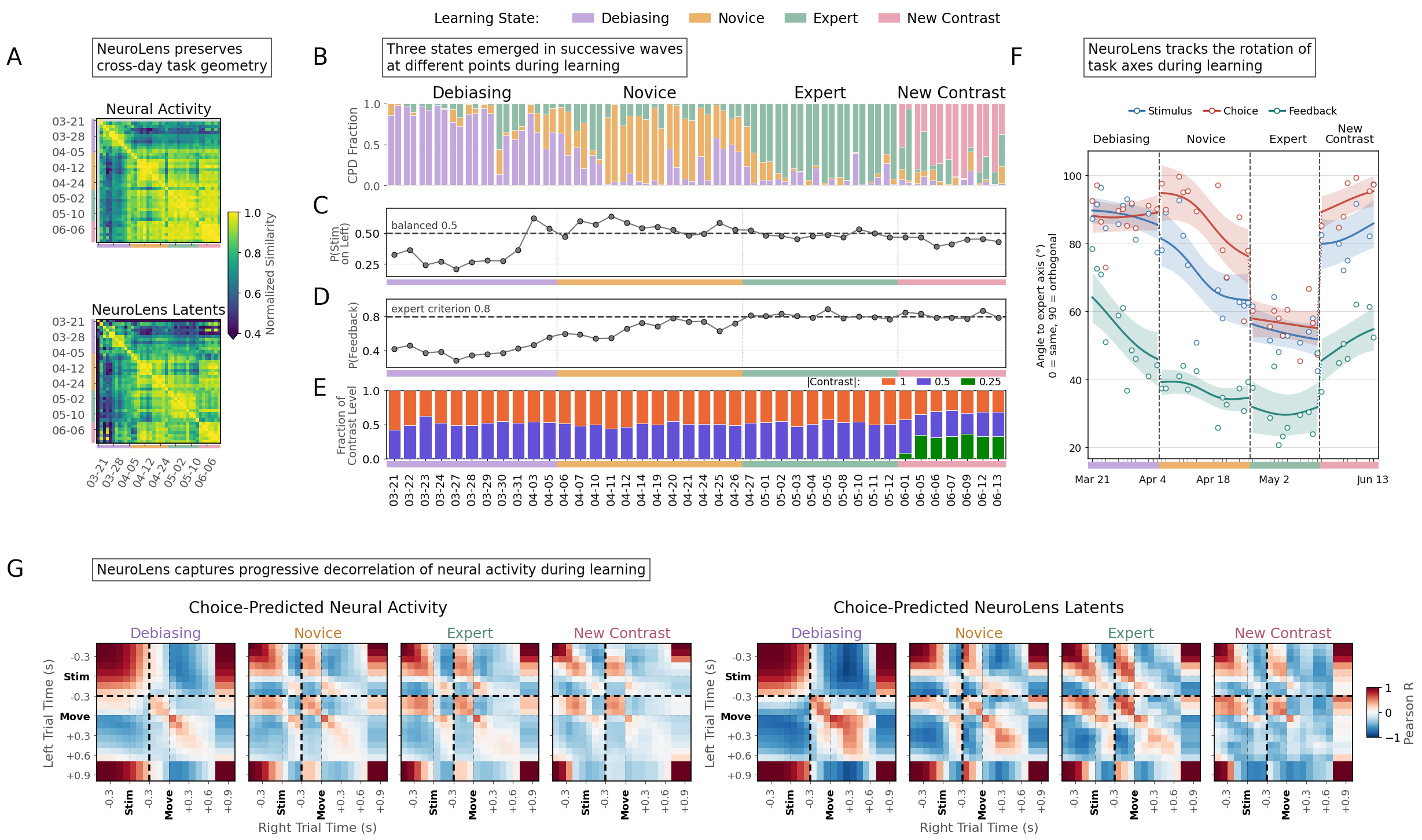}
    \vspace{-4.5mm}
    \caption{\footnotesize \textbf{{\figname} tracks and enhances gradual, task-related changes in neural activity during learning.} (A) Cross-day similarity, normalized to account for within-day variance, reveals the four learning stages in raw neural activity. {\figname} latents preserve this state-dependent structure. (B--E) Learning progresses through four behavioral states: \textit{debiasing}, characterized by biased choices and corrective stimulus sampling; \textit{novice}, when stimulus-guided decisions emerge; and \textit{expert}, when choice accuracy reaches 80\%, \textit{new contrast}, when a new, low contrast level is introduced to increase task difficulty. Coefficient of partial determination (CPD) analysis of wheel kinematics reveals successive, overlapping contributions of these states (B); stimulus-location distributions (C), choice accuracy (D) and stimulus contrast level (E) define the learning-state boundaries. (F) {\figname} tracks the rotation of stimulus, choice and feedback axes in latent space during learning. Each task axis is the vector connecting the average latent representations of two task conditions within a specified time window, and its angle is measured relative to the corresponding mean expert-stage axis. Stimulus and choice axes reorganize most rapidly during the novice stage, when the animal is initially learning the task, and stabilize during expert performance (a non-learning phase), whereas feedback-axis reorganization begins during debiasing. During the new contrast stage, all three task axes reorganize as the animal relearns to adapt to changing task demands. Circles denote sessions; curves show Gaussian-kernel smooths within each learning state with 95\% confidence intervals. (G) Choice-predicted neural population patterns for left and right choices become progressively decorrelated during learning, a progression preserved in {\figname} latents.} \label{fig:figure4}
\end{figure}

\paragraph{Neural and behavioral changes reveal four learning stages.} To identify learning stages in the dataset, we compare neural population patterns across days. Cross-day similarity in neural activity, normalized to account for within-day variance, reveals four clusters of neural states. {\name} latents preserve this stage-dependent structure (Figure~\ref{fig:figure4}A). We characterized these clusters using the distribution of stimulus locations and the proportion of correct choices (feedback), and stimulus contrast level as \textit{debiasing}, \textit{novice}, \textit{expert}, and \textit{new contrast}. During \textit{debiasing}, the animal favors one choice direction, and corrective stimulus sampling presents more stimuli on the opposite side to encourage balanced choices. In the \textit{novice} state, stimulus-guided decisions emerge as the animal learns to associate correct choices with reward. In the \textit{expert} state, choice accuracy reaches 80\%, indicating proficient task performance. In the \textit{new contrast} stage, a new, lower stimulus contrast level is introduced to increase task difficulty, requiring the animal to adapt its previously learned stimulus-guided behavior. To understand whether these states emerge abruptly or gradually, we use coefficient of partial determination (CPD) analysis \citep{sun2025learning} to quantify their contributions to wheel-speed patterns. These contributions appear in successive, overlapping waves (Figure~\ref{fig:figure4}B), indicating gradual behavioral transitions. 

\paragraph{Task axes reorganize on distinct learning timescales.}
To characterize learning-related neural changes, we define stimulus, choice and feedback axes as vectors connecting the average latent representations of the corresponding task conditions. We compute each latent task axis using {\name} latents from the corresponding task time windows: the first 100~ms after stimulus onset for stimulus, the 100~ms before movement onset for choice, and the first 200~ms after feedback onset for feedback \citep{international2025brain}. We then measure each axis's angle relative to its mean direction across expert-stage sessions. Stimulus and choice axes reorganize most rapidly during the novice stage and stabilize during expert performance, whereas feedback-axis reorganization begins during debiasing. During the new-contrast stage, all three task axes reorganize as the animal relearns to adapt to changing task demands (Figure~\ref{fig:figure4}F). These trajectories indicate that task-associated directions in latent space approach their expert-stage orientations on different timescales and reflect a reorganization of neural activity toward task-relevant directions. 

\paragraph{Latents preserve progressive decorrelation of choice-related activity.} We fit encoding models to predict neural activity or latents from stimulus location and contrast, choice, feedback, wheel speed, and other movement variables. To isolate choice-related structure, we compute correlations between population vectors from left- and right-choice trials using choice-predicted neural activity or latents, aligned to stimulus onset and first movement onset \citep{sun2025learning}. Figure~\ref{fig:figure4}G shows that neural population patterns for left and right choices become progressively decorrelated as learning advances, suggesting increasingly orthogonalized choice-related representations. {\name} latents preserve this progression (Figure~\ref{fig:figure4}G). 

Together, these findings suggest that {\name} tracks gradual, task-related changes in neural activity during learning. However, they do not establish full disentanglement of learning from recording drift, as the observed changes may also reflect unmeasured task variables and recording-related variability.

\par
\endgroup

\section{Discussion}
In this work, we introduce {\name}, a JEPA-based self-supervised model that learns denoised, semantically informative representations from chronic neural recordings. {Across mouse and human intracortical data, it learns task-relevant latent geometry that supports decoding of high-level task variables, maintains temporal consistency that supports stable decoding across days, and enables rapid few-shot adaptation to unseen neural populations. On the FALCON benchmark's M2 track, {\name} ranks first among non-ensemble methods across all three cross-day adaptation settings, more than doubling the strongest competing zero-shot baseline's held-out decoding performance. These results demonstrate consistent generalization across recording days, even without calibration data. Beyond supporting stable decoding, its latents preserve task-related representational changes, and can be used to study how neural latents evolve during learning. However, the method is not explicitly designed to disentangle learning- and behavior-related changes from recording instability. We also do not model cross-area interactions or region-specific representational changes. Future work combining more comprehensive behavioral modeling with cross-area analyses could help distinguish these sources of variability.}

\subsection*{AI Use Statement}

The authors used generative AI tools for limited assistance with model implementation and debugging, as well as manuscript refinement. All AI-assisted content was reviewed by the authors, who take responsibility for the final work.

\bibliography{iclr2027_conference}
\bibliographystyle{iclr2027_conference}


\appendix

\input{appendix/calibration}


\input{appendix/data_preprocessing}


\input{appendix/data_split}


\input{appendix/metrics}


\input{appendix/pretrain_finetune}


\input{appendix/falcon_M2}


\input{appendix/ablation}


\input{appendix/efficiency}

\input{appendix/2-stage_training}

\input{appendix/learning}

\input{appendix/model}


\input{appendix/training}

\end{document}

%% file: math_commands.tex
\usepackage{amsmath,amsfonts,bm}

\def\eqref#1{equation~\ref{#1}}

\def\1{\bm{1}}

\DeclareMathAlphabet{\mathsfit}{\encodingdefault}{\sfdefault}{m}{sl}
\SetMathAlphabet{\mathsfit}{bold}{\encodingdefault}{\sfdefault}{bx}{n}



%% file: appendix/calibration.tex
\section{Statistical Neuron ID Embeddings vs. Per-Day FiLM Drift Adaptation}\label{app:stats_vs_film}

We use statistical neuron-ID embeddings (Appendix~\ref{app:neuron_signatures}) for single-unit data, such as IBL’s chronic Neuropixels recordings, and per-day FiLM (Appendix~\ref{app:channel_drift_stats}) for multi-unit data, such as Utah array recordings. Both statistical neuron-ID embeddings and per-day FiLM support adaptation to unseen neural populations, but they address different sources of variability: statistical neuron IDs accommodate changes in the recorded neurons through matching their functional similarity, whereas FiLM infers and compensates for day-specific drift.

Mechanistically, statistical neuron ID embeddings leverage the firing properties of neurons (FR, ACG, and PC) to capture functional similarity that can transfer across sessions without knowing a neuron’s fixed identity. Once we embed these neurons’ firing signatures into the same latent space, similarly behaving neurons are mapped to similar locations in the token space. We do not use this approach for multi-unit Utah array data because (1) each channel records from multiple units, mixing their firing properties and making their signatures less reliable than those of single units; (2) it performs poorly empirically on Utah array data (Appendix~\ref{app:ablation_channel_adaptation}); and (3) prior work on Utah array-based human BCI recordings commonly models per-day drift using day-specific affine transformations \citep{willett2023high, fan2023plug, card2024accurate}.

Therefore, we use per-day FiLM, a nonlinear extension of a per-day affine transformation, for multi-unit Utah array data. The MLP in per-day FiLM contains multiple linear layers, each with a hypernetwork trained to predict the parameters of each linear layer. When the model encounters drift on a new day, it can infer the parameters needed to compensate for the drift based on patterns learned during training.

\par

\section{Calibration-Based Statistical Neuron Identification}\label{app:neuron_signatures}

For each recording day \(d\), we compute a fixed physiological and functional signature for every neuron using only a small set of calibration trials. Let
$
\mathbf{x}^{(d,k)} \in \mathbb{R}^{T \times N_d}, k=1,\ldots,K,
$
denote the binned spike-count matrix from calibration trial \(k\) on day \(d\), where \(N_d\) is the number of recorded neurons and \(x^{(d,k)}_{t,n}\) is the spike count of neuron \(n\) in time bin \(t\). By default, \(K=64\) training trials are selected once and held fixed. The resulting signature \(\mathbf{s}^{(d)}_n\) is computed without gradients and frozen during model training. As described in the main text, the learnable network \(h_{\mathrm{id}}\) maps this signature to the neuron embedding \(\mathbf{e}^{(d)}_n\). We compute the following statistics independently for each calibration trial and then average them across trials, and use $\epsilon = 10^{-6}$ for numerical stability.

\paragraph{Firing Rate Statistics.}
For each neuron \(n\) and calibration trial \(k\), we define the mean and variance as follows:
$$
\mu^{(d,k)}_n
=
\frac{1}{T}\sum_{t=1}^{T}x^{(d,k)}_{t,n},
\qquad
\left(\sigma^{(d,k)}_n\right)^2
=
\frac{1}{T-1}\sum_{t=1}^{T}
\left(x^{(d,k)}_{t,n}-\mu^{(d,k)}_n\right)^2.
$$
The firing-rate-based neuron signature is computed as follows:
{\normalsize
$$
\mathbf{s}^{(d)}_{n,\mathrm{FR}}
=
\left\langle
\left[
\log\!\left(1+\mu^{(d,k)}_n\right),\;
\log\!\left(1+\sigma^{(d,k)}_n\right),\;
\frac{\left(\sigma^{(d,k)}_n\right)^2}{\mu^{(d,k)}_n+\epsilon},\;
\frac{\sigma^{(d,k)}_n}{\mu^{(d,k)}_n+\epsilon},\;
\frac{1}{T}\sum_{t=1}^{T}
\mathbf{1}\{x^{(d,k)}_{t,n}>0\}
\right]
\right\rangle_{k},
$$
}
where \(\langle\cdot\rangle_k\) denotes averaging over the \(K\) calibration trials. These five statistics are the log mean rate, log standard deviation, Fano factor \citep{churchland2010stimulus}, coefficient of variation, and fraction of active time bins.

\paragraph{Autocorrelogram.}
Let
$
\widetilde{x}^{(d,k)}_{t,n}
=
x^{(d,k)}_{t,n}-\mu^{(d,k)}_n
$
be the mean-centered spike count. For lags \(l=1,\ldots,L\), we define the $l$-lag autocorrelation-based neuron signature $\mathbf{s}^{(d)}_{n,\mathrm{ACG}}
\in\mathbb{R}^{L}$ as follows:
$$
\mathbf{s}^{(d)}_{n,\mathrm{ACG}}(l)
=
\left\langle
\frac{
\frac{1}{T-l}
\sum_{t=1}^{T-l}
\widetilde{x}^{(d,k)}_{t,n}
\widetilde{x}^{(d,k)}_{t+l,n}
}{
\frac{1}{T}\sum_{t=1}^{T}
\left(\widetilde{x}^{(d,k)}_{t,n}\right)^2+\epsilon
}
\right\rangle_k.
$$
This normalized autocorrelogram captures each neuron’s intrinsic timescale and its tendency toward bursting or refractoriness. We omit the 0-lag term because normalization makes it approximately one by construction. For the IBL data, we use $L = 20$ lags.

\paragraph{Population Coupling.}
Within each calibration trial $k$, we z-score each neuron's spike counts:
$
z^{(d,k)}_{t,n}
=
\frac{\widetilde{x}^{(d,k)}_{t,n}}
{\sigma^{(d,k)}_n+\epsilon}.
$
We then define the leave-one-neuron-out population signal as
$
p^{(d,k)}_{t,n}
=
\sum_{\substack{j=1\\j\neq n}}^{N_d}
z^{(d,k)}_{t,j}
$
and normalize it to obtain $\overline{p}^{(d,k)}_{t,n}$. For lags \(l=0,\ldots,L-1\), the $l$-lag population-coupling-based neuron signature $\mathbf{s}^{(d)}_{n,\mathrm{PC}}
\in\mathbb{R}^{L}$ is defined as follows:
$$
\mathbf{s}^{(d)}_{n,\mathrm{PC}}(l)
=
\left\langle
\frac{1}{T-l}
\sum_{t=1}^{T-l}
z^{(d,k)}_{t,n}\,
\overline{p}^{(d,k)}_{t+l,n}
\right\rangle_k.
$$
This is the Pearson cross-correlation between neuron \(n\) and the activity of the remaining population, with the population shifted by \(l\) bins. , Positive lags indicate whether activity in neuron $n$ precedes activity in the rest of the population. Since the population signal is formed by summing over all other neurons, the feature does not depend on neuron order and can be computed for data with different numbers of neurons. For the IBL data, we use $L = 16$ lags.

\section{Calibration-Based Channel Drift Adaptation}\label{app:channel_drift_stats}

\paragraph{Channel Statistics.} For each recording day $d$, we use a fixed set of calibration trials to compute five summary statistics for each channel and then average these statistics across trials. The choice of statistics depends on the input data. For T12 spike count data \citep{willett2023high}, which are non-negative and count-valued, we use the firing-rate statistics described in Appendix~\ref{app:neuron_signatures}: mean activity, standard deviation, Fano factor \citep{churchland2010stimulus}, coefficient of variation, and the fraction of nonzero time bins. In contrast, spike-band power (SBP) is not count-valued, and the T15 dataset provides pre-z-scored features without raw spike counts \citep{card2024accurate}. For these inputs, we instead use statistics that describe the distribution's location, spread, skewness, tail weight, and magnitude. 

\paragraph{Per-Day Drift Context Vector.} We normalize each statistic using constants estimated from the training days, so that no individual statistic dominates because of its numerical scale. For each channel, we first convert its five normalized calibration statistics into a learned channel embedding using the same MLP layer. This shared mapping lets the model represent different patterns of activity and recording drift in a common feature space. We then average the channel embeddings separately within each input modality (spike counts and SBP). The resulting modality-level vectors summarize the typical calibration distribution for that modality on a given day. Finally, we concatenate these modality summaries and project them into a single day context vector $\mathbf{c}^{(d)}$. Because the channel embeddings are averaged, the context vector $\mathbf{c}^{(d)}$ does not depend on the order of channels.

\paragraph{Per-Day FiLM Conditioning} Conditioned on $\mathbf{c}^{(d)}$, a shared hypernetwork generates the scale and shift parameters of a day-specific feature-wise linear modulation (FiLM) layer \citep{perez2018film}:
$$
\left(\boldsymbol{\gamma}^{(d)}_\ell,
      \boldsymbol{\beta}^{(d)}_\ell\right)
=
h_{\mathrm{hyper},\ell}\!\left(\mathbf{c}^{(d)}\right), \qquad
\widetilde{\mathbf{h}}^{(d)}_{t,\ell}
=
\left(1+\boldsymbol{\gamma}^{(d)}_\ell\right)
\odot \mathbf{h}^{(d)}_{t,\ell}
+
\boldsymbol{\beta}^{(d)}_\ell,
\label{eq:film}
$$
where $\mathbf{h}^{(d)}_{t,\ell}$ is the output at time $t$ in encoder layer $\ell$. The FiLM layer therefore applies a day-specific, feature-wise affine transformation to the encoder layers. The hypernetwork parameters are shared across days and learned during training. For an unseen day, calibration trials are used only to compute $\mathbf{c}^{(d)}$, which generates the corresponding FiLM parameters without updating the model weights.

%% file: appendix/data_preprocessing.tex
\section{Data Preprocessing}\label{app:data_preprocessing}

\subsection{IBL chronic Neuropixels data}\label{app:ibl_data}

For the IBL chronic Neuropixels dataset, we analyzed neural and video recordings collected over 3 months from a single mouse (\textit{MM008}) \citep{melin2024large}. We derived binned spike counts from the neural recordings and behavioral variables from the video recordings, as described below.

\paragraph{Neural Activity (\texttt{spikes}).} For each recording session, we combined spike-sorted units using \textit{Kilosort} \citep{pachitariu2016kilosort} from all probes into a single neural population. We retained units with a mean firing rate greater than 1 Hz, computed using only the training portion of each recording; thus, unit selection did not use validation or test data. Recorded units were located mainly in primary and secondary motor cortex and dorsal striatum, with smaller numbers in prelimbic and dorsal anterior cingulate cortex. Spike times were binned into 16.7 ms intervals ($\sim$60 Hz), and raw spike counts were used as model inputs.

We divided each session into trial and non-trial periods. For \textit{trial} data, each trial was represented by a 2 s window spanning 0.8 s before to 1.2 s after stimulus onset (120 bins). For \textit{non-trial} data, we excluded the interval from 0.2 s before to 0.8 s after stimulus onset for every valid trial. We segmented the remaining recording into consecutive, non-overlapping 2 s windows (120 bins).

\paragraph{Trial-Level Behavior.} We used the mouse's binary choice in the visual decision-making task as a discrete behavioral variable. On each trial, the mouse reported whether it perceived the stimulus on the left or right side. Choice labels were therefore available only for trial windows; non-trial windows were unlabeled.

\paragraph{Time-Bin-Level Behavior.} From video recordings, we extracted wheel speed, whisker motion energy from the left and right cameras, and paw and nose speeds from the left camera using keypoint estimation \citep{biderman2024lightning}. Each signal was resampled to the same 16.7 ms bins used for spike counts. Because these measures are available during both trial and non-trial periods, we included them in both window types. Frames with missing keypoints were assigned a speed of zero and were not interpolated. Each behavioral variable was scaled to the range [0, 10] using its minimum and maximum values from the training portion of the recording only. The resulting scaling parameters were then applied consistently across all data splits and window types.

\subsection{Human chronic Utah array data.}

\paragraph{Neural Activity (\texttt{spikes} and \texttt{SBP}).} We used channel-level features rather than spike-sorted units. Both T12 \citep{willett2023high} and T15 \citep{card2024accurate} provide threshold-crossing counts and spike-band power (SBP) for each electrode in 20 ms bins. We smoothed each feature separately within each trial using a Gaussian kernel with a standard deviation of 40 ms, then combined every five consecutive bins into a 100 ms input token. Because the sentences differ in length, trials have variable durations.

For T12, we used the 128 electrodes in ventral premotor cortex (area 6v) and excluded electrodes in area 44. The two features from each electrode yielded 256 input dimensions per time bin. We first z-scored each feature within each recording session, then z-scored it again within each day using statistics computed only from that day's training trials. For T15, we used all 256 electrodes in the precentral gyrus, which yielded 512 input dimensions per time bin. We used the publicly released, normalized T15 features without further normalization.

\paragraph{Phoneme Labels.} We derived phoneme labels following the procedures of \citet{willett2023high, card2024accurate}. For T12, we converted each sentence to a phoneme sequence with the \textit{g2p-en} grapheme-to-phoneme package. Before conversion, we lowercased the text and removed punctuation other than apostrophes. We then removed stress markers and added a silence token at word boundaries and at the end of each sentence. This produced a vocabulary of 39 phonemes and one silence token; we added a blank token for CTC training. For T15, we used the phoneme sequences provided with the dataset, which use the same 40 non-blank classes. When training supervised baselines jointly across T12 and T15 datasets, we remapped T15 labels to the T12 class order to ensure consistent labeling across participants.

\paragraph{Sentence Embeddings.} We encoded each sentence with the pretrained \texttt{all-MiniLM-L6-v2} model \citep{reimers2019sentence}. We averaged its contextual token embeddings and L2-normalized the result to obtain one 384-dimensional vector per sentence. The shared encoder placed sentence representations from both participants in the same embedding space.

%% file: appendix/data_split.tex
\section{Data Splits}
\label{app:data}

For both the IBL and human chronic datasets, we considered two experimental settings: (1) \textit{high- vs.\ low-level neural decoding} (Section~\ref{sec:high_vs_low_decoding}) and (2) \textit{generalization across time} (Section~\ref{sec:across_time}). The two settings use the same model architectures and training procedures (Appendices~\ref{app:model_details} and~\ref{app:training_details}) and differ only in how the data are split. In the first setting, training and test data were drawn from the same recording days, allowing us to evaluate representation quality without requiring generalization across days. In the second setting, we split the data chronologically: models were trained on earlier recording days and evaluated on later, held-out days, creating a temporal distribution shift between training and test data.

\paragraph{IBL chronic Neuropixels data.} For \textit{high- vs.\ low-level neural decoding}, we focused on single-session models. Within each session, we split the data into 70\% training, 10\% validation and 20\% test sets. Because all splits came from the same recording day, this setting does not introduce a substantial temporal distribution shift. For \textit{generalization across time}, we split the full \textit{MM008} dataset chronologically to introduce an across-day distribution shift. We used 39 days collected between \textit{2023-03-21} and \textit{2023-05-17} for multi-day pretraining, and held out 14 subsequent days, collected between \textit{2023-05-18} and \textit{2023-06-13} for evaluation. When adapting the model to each held-out day, we used the first 20\% of trials as the adaptation set, the next 10\% as the validation set for checkpoint selection, and the remaining 70\% to evaluate decoding performance.

\paragraph{Human chronic Utah array data.} For \textit{high- vs.\ low-level neural decoding}, all data splits were drawn from the same recording days. For T12 \citep{willett2023high}, we use the provided
training partition for training, hold out the last block of each
training day for validation, and use the provided evaluation
partition as the test set, resulting in 7,920 training, 880
validation, and 880 test trials across 24 recording days. For T15
\citep{card2024accurate}, we use the provided training partition and
divide the provided evaluation partition equally into validation and
test sets, resulting in 8,072 training, 704 validation, and 722 test
trials across 45 days. For \textit{generalization across time}, we split each participant’s full dataset chronologically, using the earliest 80\% of recording days for training and holding out the final 20\% for evaluation. For T12, we trained on 18 past days and evaluated on the subsequent 5 days (excluding
2022-06-14 due to data quality). For T15, we trained on 36 past days and evaluated on the following 9 days. Within each training day, we used the first 90\% for training and the remaining 10\% for validation. For each held-out day, the first 20\% served only as unlabeled calibration data, and we reported decoding performance on the remaining 80\%.

%% file: appendix/metrics.tex
\section{Metrics}
\subsection{Effective Rank of Neural Latents}\label{app:latent_rank}

We use effective rank to quantify the dimensionality of a latent representation. Intuitively, it measures the number of dimensions that contribute substantially to the representation. For each recording day, we estimated the within-trial covariance $\Sigma$ across the trial-aligned time windows (Appendix~\ref{app:data_preprocessing}). We sampled the same number of trials from each day so that differences in the computed effective rank were not driven by differences in trial count. For a covariance matrix $\Sigma$ with eigenvalues $\lambda_1,\ldots,\lambda_D \geq 0$, where $D$ is the latent dimension, let
$
p_i = \frac{\lambda_i}{\sum_{j=1}^{D} \lambda_j}.
$
The effective rank is the exponential of the Shannon entropy of this normalized eigenvalue spectrum: \begin{equation*}
\mathrm{erank}(\Sigma)
= \exp \left(
-\sum_{i=1}^{D} p_i \log p_i
\right),
\label{eq}
\end{equation*}
which ranges from 1, when variance is concentrated in a single direction (representation collapse), to $D$, when variance is distributed equally across all dimensions. We computed the effective rank of the PCs of raw neural activity $\mathbf{x}^{(d)}_t$, the causal NDT neural latents, and the {\name} latents $\mathbf{\hat{z}}^{(d)}_t$. Figure~\ref{fig:figure4} shows how these values change across recording days.

\subsection{Phoneme Error Rate}\label{app:per}
We used a 41-token phoneme vocabulary comprising all phonemes, silence, and the CTC blank token. We computed the phoneme error count by first removing repeated tokens and blanks from the decoded phoneme sequences and then calculating the Levenshtein edit distance \citep{levenshtein1966binary} between the decoded and ground-truth sequences, counting substitutions, insertions, and deletions. We then normalized the error count by the total number of ground-truth phonemes to obtain the phoneme error rate.

\subsection{Top-5 Sentence Retrieval Accuracy}\label{app:sentence_retrieval}
Top-5 sentence retrieval accuracy measures whether a model's predicted sentence embedding is close enough to the ground-truth (GT) sentence to pick it out of a candidate pool: a trial counts as correct if the GT sentence is among the 5 most similar to the prediction. We use this metric on the T12 and T15 datasets to evaluate representation quality via the semantic decoding task, where the target is the 384-dimensional \texttt{all-MiniLM-L6-v2} sentence embedding \citep{reimers2019sentence}. We report average top-5 accuracy across held-out days. Each day has a candidate pool of 288 $\sim$ 448 sentences, which keeps chance level low (1$\sim$2\%). Predicted and GT candidate embeddings are first centered by subtracting the mean GT embedding across test trials, so we can focus on trial-to-trial variability; candidates are then ranked by cosine similarity to the centered prediction.

\subsection{Representation Similarity Analysis}\label{app:rsa}

\paragraph{Representational similarity analysis.} Representational Similarity Analysis (RSA) \citep{kriegeskorte2008representational} compares representations through their second-order structure. Rather than aligning representations point by point, RSA first converts each representation into a representational dissimilarity matrix (RDM), which quantifies the dissimilarity between every pair of conditions (time bins within a trial). It then compares two representations by correlating their RDMs. Because this comparison depends on pairwise relationships rather than the coordinate system itself, RSA is invariant to rotations of the underlying representation and can be applied across models or recording days without a shared basis.

\paragraph{Cross-day representational dissimilarity.} For each day $d$ and representation (PCs of raw neural activity $\mathbf{x}^{(d)}_t$, causal NDT latents, or {\name} latents $\widehat{\mathbf{z}}^{(d)}_t$), we constructed a temporal RDM, $\mathrm{RDM}^{(d)} = \mathbf{1} - \mathbf{P}^{(d)}$, where $\mathbf{P}^{(d)} \in \mathbb{R}^{T \times T}$ contains Pearson correlations between time bins. For a day pair $d_i$ and $d_j$, we quantified their dissimilarity as
$$
\tau_{i,j} = 
1-\rho_\text{spearman}\left(
\mathrm{RDM}^{(d_i)},\,
\mathrm{RDM}^{(d_j)}
\right),
$$
where $\rho_\text{spearman}$ denotes the Spearman correlation. This measure is zero when the two days have identical temporal geometries and increases as their geometries diverge. 

\paragraph{Within-day correction for sampling variability.} To adjust for within-day sampling variability, we estimated the within-day split-half drift ($\tau_{i,i}$), by computing RSA between RDMs estimated from two randomly selected halves of trials from day $d_i$. Our primary measure was the within-day-drift-corrected dissimilarity:
$$
\widetilde{\tau}_{i,j}
= \tau_{i,j} - \frac{1}{2}\left(
\tau_{i,i}
+
\tau_{j,j}
\right).
$$
We visualized $\widetilde{\tau}_{i,j}$ as a function of the lag between each pair of days in Figure~\ref{fig:figure4}, which provides a model-free characterization of representational drift.

%% file: appendix/pretrain_finetune.tex
\begin{figure}[h!]
    \centering
    \includegraphics[width=\textwidth]{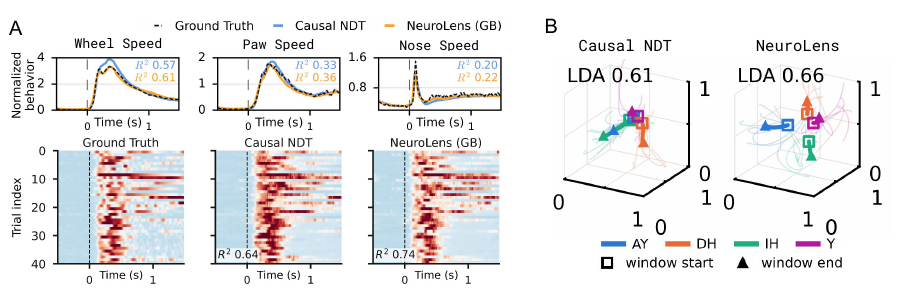}
    \vspace{-4mm}
    \caption{\footnotesize \textbf{{\figname} captures trial-to-trial behavioral variability and enhances phoneme separation across held-out days.} (A) Comparison of predicted movement-related behavior averaged across all trials (top) and predicted behavior from single trials (bottom), shown to demonstrate that the baselines can predict trial-to-trial variability in behaviors. The $R^2$ values for both trial-average and single-trial behavior are shown. The dashed line shows the stimulus onset. (B) Latent trajectories for different phoneme onsets, projected onto the first three PCs, across held-out days in T15. Dark (light) lines show averaged (individual) time windows. {\figname} shows clearer phoneme separation than causal NDT, indicating better held-out adaptation, quantified by LDA balanced classification accuracy.} \label{fig:decoding_quality}
\end{figure}

\section{Pretraining}\label{app:pretraining}

Multi-day pretraining on prior sessions is beneficial only if it outperforms training directly on the target session. We evaluated this on the IBL dataset while holding the target-session data budget fixed: a single-session model was trained from scratch on the same training split used to adapt the pretrained model. Thus, the two baselines differed only in exposure to the 39 prior sessions, with identical model architectures and training procedures.

\paragraph{Setup.} We randomly selected 5 of the 14 held-out sessions using a fixed seed. For each session, a {\name} model was trained from random initialization on the session's 20\% adaptation split, selected using the same 10\% validation split, and evaluated on the same held-out 70\% test split as its pretrained counterpart (Appendix~\ref{app:data}). The architecture, optimizer, learning-rate schedule, and epoch budget matched those of the pretrained counterpart (Appendix~\ref{app:training_details}); the two models were parameter-matched. 

\paragraph{Result.} In Table~\ref{tab:pretraining_ablation}, pretraining improves movement-related behavior decoding on every session tested by 0.046 to 0.094 $R^2$ (mean 0.076), which is large relative to the spread across baselines in Section~\ref{sec:across_time}. For choice decoding, the pretrained model is ahead on four of the five sessions.

\begin{table}[h!]
\caption{Pretraining on 39 days vs. training on the target
day alone, on 5 held-out sessions drawn at random. Both baselines under comparison use the same 20\% / 10\% / 70\% adaptation/validation/test split for few-shot finetuning and the same architecture; they differ in whether {\name} is initialized with a 39-day pretrained backbone. Movement behavior $R^2$ is the mean over the 5 time-bin-level variables.}
\begin{center}
\small
\begin{tabular}{lcccc}
\toprule
 & \multicolumn{2}{c}{Movement Behavior ($R^2$) $\uparrow$} & \multicolumn{2}{c}{Choice (Balanced Accuracy) $\uparrow$} \\
\cmidrule(lr){2-3}\cmidrule(lr){4-5}
Session & Single-day & 39-day pretrained & Single-day & 39-day pretrained \\
\midrule
2023-05-18 & 0.42 & \textbf{0.47} & 0.72 & \textbf{0.76} \\
2023-05-22 & 0.41 & \textbf{0.48} & 0.63 & \textbf{0.76} \\
2023-05-24 & 0.35 & \textbf{0.44} & 0.83 & 0.61 \\
2023-05-30 & 0.38 & \textbf{0.47} & 0.63 & \textbf{0.69} \\
2023-06-09 & 0.33 & \textbf{0.42} & 0.67 & \textbf{0.82} \\
\midrule
Mean & 0.38 & \textbf{0.45} & 0.69 & \textbf{0.73} \\
\bottomrule
\end{tabular}
\end{center}
\label{tab:pretraining_ablation}
\end{table}

\section{Phoneme and Semantic Decoding}\label{app:phoneme_semantic_decoding}

For each baseline, once its Transformer-, GRU-, or MLP-based backbone was pretrained, we attached a task-specific decoder and fine-tuned the backbone end-to-end to decode the target. For pretraining, data from T12 and T15 were pooled; the resulting model was then fine-tuned separately for phoneme and sentence-embedding decoding.

\paragraph{Phoneme decoding.} A linear decoder head mapped the sequence of backbone outputs (e.g., $\widehat{\mathbf{z}}_{1:T}^{(d)}$ in Eq.~\ref{eq:ar_predictor}) to phoneme logits, including the CTC blank token. We trained the model with CTC loss, which does not require time-bin-level alignment between neural activity and phoneme labels. At inference, we applied standard CTC decoding and evaluated performance using phoneme error rate (PER; Appendix~\ref{app:per}).

\paragraph{Semantic decoding.} We pooled the sequence of backbone outputs (e.g., $\widehat{\mathbf{z}}_{1:T}^{(d)}$ in Eq.~\ref{eq:ar_predictor}) by averaging across the $T$ time bins to obtain a single sentence token, which we mapped to the target sentence embedding space with a task-specific decoder head. The decoder was trained with a cosine similarity objective to encourage alignment in representation space. At inference, we compared predicted embeddings with candidate sentence embeddings from a closed-book vocabulary and ranked candidates by cosine similarity. We report top-5 sentence retrieval accuracy; higher values indicate more accurate sentence retrieval and, therefore, better semantic decoding.

%% file: appendix/falcon_M2.tex
\section{FALCON benchmark M2 Track}\label{app:falcon_M2}

\begingroup

\paragraph{Monkey chronic Utah array data and benchmark evaluation (FALCON M2).}
The M2 task of the FALCON benchmark \citep{karpowicz2024falcon} evaluates whether neural decoders trained on earlier recording days generalize to recordings collected days to weeks later. A rhesus macaque moved its fingers to acquire cued targets while spiking activity was recorded from 96-channel Utah arrays in the precentral gyrus across 13 sessions over 8 recording days spanning 36 days. Neural features are threshold-crossing spike counts binned at 20\,ms, and the decoding target is 2-D finger velocity. The first 4 days (7 sessions) are held-in, and the last 4 days (6 sessions) are held-out; three held-out days occurred 3$\sim$4 weeks after the last held-in day.

Trials within each session are split chronologically. For held-in sessions, the first 60\% is released with labels for training (1,923 trials, $\sim$39 minutes in total). For held-out sessions, only the first 10\% is released for calibration, with its use determined by the evaluation setting (Table~\ref{tab:falcon_settings}). Submitted decoders operate causally, receiving one 20-ms bin at a time without labels or trial boundaries, and are scored on a private test split comprising the last 40\% of each session. The ranking metric is variance-weighted $R^2$, computed per held-out day and averaged across the four held-out days; performance on held-in days is reported for reference. 

\par
\endgroup

\begingroup

\paragraph{Adaptation of {\name} to FALCON benchmark.}
To match FALCON's causal streaming evaluation, the model is trained on 50-bin windows and decodes the velocity of the last bin of each window. A linear decoder reads the concatenation of the encoder embedding at the current bin and the predictor's forecast embedding for that bin. The settings in Table~\ref{tab:falcon_settings} differ in how held-out data are used.

In the \textit{zero-shot (ZS)} setting, {\name} is trained exclusively on held-in data, with no access to held-out data. In the \textit{gradient-free few-shot unsupervised (GF-FSU)} setting, {\name} uses the gradient-free adaptation: each session's neural inputs are z-scored per channel using statistics computed from that session's calibration spikes. Adaptation to a held-out session therefore requires only unlabeled spikes and no parameter updates. In the \textit{few-shot supervised (FSS)} setting, {\name} uses the gradient-based adaptation described in two stages. First, the GF-FSU model is fine-tuned jointly on held-in data and labeled calibration trials from all held-out sessions. Then, for each held-out session, only the read-in layer is further fine-tuned on that session's labeled calibration trials.

\par
\endgroup

\begin{table}[t!]
\centering
\small
\caption{FALCON M2 evaluation setup. $^{*}$ denotes reference decoders trained by the FALCON benchmark hosts using private calibration trials unavailable to benchmark participants. These decoders approximate an upper bound on decoding performance.}
\label{tab:falcon_settings}
\begin{tabular}{@{}llcc@{}}
\toprule
Cross-day adaptation & Fraction of held-out data used & Labels provided & Model parameter updates \\
\midrule
Zero-shot (ZS) & None & None & $\times$ \\
Few-shot unsupervised (FSU) & First 10\% & $\times$ & \checkmark \\
Gradient-free FSU (GF FSU) & First 10\% & $\times$ & $\times$ \\
Few-shot supervised (FSS) & First 10\% & \checkmark & \checkmark \\
Oracle $^{*}$ & First 60\% & \checkmark & \checkmark \\
\bottomrule
\end{tabular}
\end{table}

\begin{table}[t!]
\centering
\small
\setlength{\tabcolsep}{4pt}
\caption{Training hyperparameters on FALCON M2. All models listed in the table use 50-bin windows with a stride of 1 and an MSE loss on finger velocity at the final time bin, weighted by 5. For {\name}, we additionally apply the self-supervised loss described in Section~4.2, weighted by 0.1. All models use the Adam optimizer without a learning-rate scheduler, a batch size of 32, and dynamic channel dropout. {\name} and SPINT \citep{le2026spint} have 4.68\,M and 4.59\,M parameters, respectively. FSS models are retrained on all calibration trials for the selected number of epochs.}
\label{tab:falcon_hparams}
\begin{tabular}{@{}>{\raggedright\arraybackslash}p{2.3cm}
                *{3}{>{\raggedright\arraybackslash}p{2.1cm}}
                >{\raggedright\arraybackslash}p{3.9cm}@{}}
\toprule
 & SPINT \citep{le2026spint}
 & {\name} (ZS)
 & {\name} (GF-FSU)
 & {\name} (FSS: stage 1 $\rightarrow$ stage 2) \\
\midrule
Model initialization
 & Random & Random & Random & GF-FSU model \\

\addlinespace[6pt]
Training data
 & Held-in
 & Held-in
 & Held-in
 & Held-in + calibration $\rightarrow$ calibration from each held-out session \\

\addlinespace[6pt]
Cross-day adaptation
 & Calibration-based neuron identities
 & None
 & Calibration-based z-scoring
 & Calibration-based z-scoring \\

\addlinespace[6pt]
Updated model parameters
 & All
 & All
 & All
 & All $\rightarrow$ read-in layers (0.65\%) \\

\addlinespace[6pt]
Learning rate
 & $5\times10^{-5}$
 & $5\times10^{-5}$
 & $5\times10^{-5}$
 & $5\times10^{-5}\rightarrow 1\times10^{-4}$ \\

\addlinespace[6pt]
Checkpoint selection
 & Fixed
 & Held-in cross-validation
 & Held-out calibration $R^2$
 & Calibration validation split \\
\bottomrule
\end{tabular}
\end{table}

%% file: appendix/ablation.tex
\section{Ablation Study of Neuron Identity Embeddings}\label{app:ablation_neuron_signatures}



\begingroup

The IBL recordings contain different numbers and sets of spike-sorted neurons across sessions. Because neuron indices are defined only for each session, embeddings learned for seen neurons in training sessions cannot be directly assigned to neurons recorded on a new day. We therefore use calibration-based neuron identification, representing each neuron by firing statistics that can also be computed in unseen neurons from a new session. These signatures characterize firing properties rather than individual neuron identities, allowing the model to generalize to new neurons with similar activity patterns. As described in Section~\ref{sec:model_architecture} and Appendix~A, they combine firing-rate statistics (FR), autocorrelograms (ACG), and population coupling (PC). Here we examine how these statistical identities affect decoding in seen and unseen neural populations and measure each component's contribution to decoding performance.

\subsection{Statistical vs. Learned Neuron Identities}
\label{app:ibl_id_comparison}

To test whether firing-based statistical neuron identities (IDs) improve generalization beyond the neurons observed during training, we compare them with learned neuron-ID embeddings and a model without neuron-identity embeddings. We pretrained {\name} on 5 earlier sessions in the IBL dataset and evaluated them on 5 later, held-out sessions. We use \textit{seen neurons} to refer to recorded units from the pretraining sessions, evaluated on held-out trials from those sessions, and \textit{unseen neurons} to refer to units from the later sessions, which were not included in pretraining. We use \textit{gradient-free (GF) adaptation} to generalize to unseen neurons on a new day. Table~\ref{tab:ibl_identity_comparison}
reports the mean movement-related behavior decoding $R^2$ and choice decoding balanced accuracy for each setting. For seen neurons, statistical and learned identities achieve similar
performance. For unseen neurons, statistical
identities perform better under GF adaptation, with
the same trend on all 5 held-out sessions. Removing
neuron-identity embeddings leads to a larger reduction in decoding
performance. These results show that statistical signatures are particularly
useful for rapid, gradient-free adaptation to unseen neural populations.

\begin{table}[H]
\centering
\caption{Comparison of neuron-identity embeddings on IBL.
Movement-related behavior decoding $R^2$ and choice balanced
accuracy are averaged over evaluation sessions
(higher is better).
Seen neurons are from pretraining sessions; unseen neurons
are from 5 held-out future sessions.}
\label{tab:ibl_identity_comparison}
\vspace{1mm}
\small
\setlength{\tabcolsep}{5pt}
\begin{tabular}{@{}lcc@{}}
\toprule
Neuron Identity & Seen Neurons & Unseen Neurons (GF) \\
\midrule
\multicolumn{3}{@{}l}{\emph{Movement-related behavior $R^2$}} \\
Statistical Neuron ID (FR + ACG + PC) & $0.38\pm 0.08$ & \textbf{$0.38\pm 0.03$} \\
Learned Neuron ID & $0.36\pm 0.08$ & $0.33\pm 0.03$ \\
No Neuron ID & $0.27\pm 0.06$ & $0.29\pm 0.03$ \\
\addlinespace[3pt]
\multicolumn{3}{@{}l}{\emph{Choice balanced accuracy}} \\
Statistical Neuron ID (FR + ACG + PC) & $0.87\pm 0.06$ & \textbf{$0.66\pm 0.05$} \\
Learned Neuron ID & $0.86\pm 0.02$ & $0.62\pm 0.03$ \\
No Neuron ID & $0.64\pm 0.02$ & $0.55\pm 0.03$ \\
\bottomrule
\end{tabular}
\end{table}

\subsection{Contribution of Individual Statistical Components}
\label{app:ibl_stats_components}

The statistical neuron signatures contain firing-rate statistics (FR),
autocorrelograms (ACG), and population coupling (PC), which describe
different aspects of a neuron's activity. To examine how each component affects decoding, we remove one component at a time while keeping the other two and repeat the IBL evaluations in the previous section. Table~\ref{tab:ibl_stats_components} reports the change in movement-related behavior decoding $R^2$ and choice decoding balanced accuracy relative to the full statistical signature; negative values indicate lower performance. Overall, removing any single component has a modest effect on decoding performance (Table~\ref{tab:ibl_stats_components}). All three components contribute to movement decoding, while their effects on choice decoding vary. Removing population coupling causes the largest mean performance drop in both tasks for unseen neurons, which shows its importance for adaptation.

\begin{table}[H]
\centering
\caption{Ablation of firing-rate statistics (FR), autocorrelograms
(ACG), and population coupling (PC) in IBL neuron identities.
Values show absolute changes relative to the full statistical
signature (mean $\pm$ one standard error across 5 sessions).
The \% notation denotes absolute changes multiplied by 100,
not relative percentage changes.
Negative values indicate lower performance.}
\label{tab:ibl_stats_components}
\vspace{1mm}
\small
\setlength{\tabcolsep}{4pt}
\begin{tabular}{@{}lcc@{}}
\toprule
Removed component & Seen neurons & Unseen neurons (GF) \\
\midrule
\multicolumn{3}{@{}l}{\emph{Movement-related behavior ($\Delta R^2$)}} \\
FR excluded & $-1.51\%\pm0.38\%$ & $-1.18\%\pm0.63\%$ \\
ACG excluded & $-1.17\%\pm0.38\%$ & $-1.51\%\pm0.41\%$ \\
PC excluded & $-1.10\%\pm0.48\%$ & $-2.86\%\pm0.51\%$ \\
\addlinespace[3pt]
\multicolumn{3}{@{}l}{\emph{Choice balanced accuracy ($\Delta$ accuracy)}} \\
FR excluded & $-5.22\%\pm4.90\%$ & $+2.74\%\pm1.13\%$ \\
ACG excluded & $-7.23\%\pm8.15\%$ & $+0.93\%\pm4.11\%$ \\
PC excluded & $-3.14\%\pm3.65\%$ & $-4.67\%\pm4.38\%$ \\
\bottomrule
\end{tabular}
\end{table}

\par
\endgroup

\begingroup
\section{Ablation Study of Channel Drift Adaptation}\label{app:ablation_channel_adaptation}

Unlike the IBL recordings, which contain varying numbers and sets of
spike-sorted neurons across sessions, human Utah array recordings have a
fixed number and ordering of channels. Each channel may capture activity
from multiple nearby neurons, and the recorded signals can change across
days despite the fixed channel configuration. We therefore use
calibration-based channel drift adaptation for BCI data, rather than
assigning identities to changing sets of individual neurons as in IBL.

As described in Appendix~B,
{\name} uses calibration statistics to estimate day-specific changes in
channel activity and generate per-day FiLM parameters. For T12 spike-count
features, we use the same five firing-rate (FR) statistics as in the IBL
neuron signature, but not its autocorrelogram (ACG) or population coupling
(PC) components. For T12 spike-band power and the pre-z-scored T15
features, we use distributional statistics suited to these continuous-valued
signals. These channel-level summaries are aggregated into a day-level
context for FiLM conditioning. We evaluate this design below.

\subsection{Comparison of Day Adaptation Methods}
\label{app:bci_adaptation_methods}

To evaluate our choice of different adaptation methods for human speech decoding datasets, we compare calibration-based per-day FiLM with two
alternative approaches on the data from T12. Unlike the IBL dataset, where the number and
identity of recorded neurons change across sessions, the human dataset
has a fixed number of channels whose activity can drift over time.
We therefore examine whether statistics-conditioned FiLM is more
effective for adapting to such channel-level drift than
(1) the statistical neuron-identity approach used for the IBL dataset, and (2) a ``linear FiLM'', where we learn a hypernetwork to generate the parameters of a per-day affine transformation. Each method is trained separately on 18 past T12 recording days and evaluated on 5 held-out future days, with the same 64 unlabeled calibration trials available for each held-out day.

\begin{table}[H]
\centering
\caption{Comparison of day adaptation methods on T12 data. Results are averaged over 5 held-out days, using the same 64 unlabeled calibration trials
per day. Phoneme error rate (PER; lower is better) and top-5 sentence
retrieval accuracy (higher is better) are reported in percentages.}
\label{tab:bci_conditioning_comparison}
\vspace{1mm}
\small
\setlength{\tabcolsep}{5pt}
\begin{tabular}{@{}lcc@{}}
\toprule
Day Adaptation Method & PER (\%) $\downarrow$ & Top-5 accuracy (\%) $\uparrow$ \\
\midrule
Per-Day FiLM (Calibration Statistics) & 38.80 & 29.80 \\
Per-Day Linear FiLM & 49.20 & 22.10 \\
Statistical Neuron ID & 65.00 & 7.90 \\
\bottomrule
\end{tabular}
\end{table}

In Table~\ref{tab:bci_conditioning_comparison}, calibration-based, per-day FiLM outperforms the model using statistical neuron IDs, as employed in the IBL chronic dataset. The per-day linear FiLM baseline also underperforms our proposed per-day FiLM approach on the T12 dataset. These results support our flexible use of different adaptation strategies for neuron-based and channel-based chronic datasets.

\subsection{Contribution of Channel Drift Adaptation}
\label{app:bci_dayfilm_ablation}

While the previous ablation compares different day adaptation
methods, we further examine whether calibration-based channel drift
adaptation itself improves decoding performance. To isolate its
contribution, we compare {\name} with and without the per-day FiLM
adaptation. Both models were pretrained on data from 18 previous days from T12 and fine-tuned on data from 5 held-out future days.

\begin{table}[H]
\centering
\caption{Ablation of calibration-based channel drift adaptation
(per-day FiLM) on T12. Phoneme error rate (PER; lower is better)
and top-5 sentence retrieval accuracy (higher is better) are
reported for 5 held-out days.}
\label{tab:ablation_dayfilm_t12_perday}
\vspace{1mm}
\small
\setlength{\tabcolsep}{5pt}
\begin{tabular}{@{}lcccc@{}}
\toprule
& \multicolumn{2}{c}{PER (Phoneme) (\%) $\downarrow$}
& \multicolumn{2}{c}{Top-5 Accuracy (Sentence) (\%) $\uparrow$} \\
\cmidrule(lr){2-3}\cmidrule(lr){4-5}
Held-out day / Per-day FiLM
& $\checkmark$ & $\times$ & $\checkmark$ & $\times$ \\
\midrule
2022-08-11 & \textbf{27.23} & 27.50 & \textbf{51.39} & 40.62 \\
2022-08-13 & \textbf{26.90} & 27.51 & \textbf{48.61} & 44.10 \\
2022-08-18 & 35.19 & \textbf{32.39} & \textbf{34.38} & 32.81 \\
2022-08-23 & \textbf{34.60} & 36.48 & \textbf{35.27} & 30.80 \\
2022-08-25 & \textbf{36.83} & 38.78 & \textbf{31.47} & 30.13 \\
\bottomrule
\end{tabular}
\end{table}

In Table~\ref{tab:ablation_dayfilm_t12_perday},
per-day FiLM improves sentence retrieval accuracy across all 5 held-out days and reduces the PER on 4 of them. Together with the results in Table~\ref{tab:bci_conditioning_comparison}, these findings support the use of our proposed per-day FiLM adaptation to improve decoding performance in datasets with neural drift.

\par
\endgroup

%% file: appendix/efficiency.tex
\begingroup

\section{Practical Data and Computational Requirements}\label{app:efficiency}

We further investigate the practical requirements of {\name} for longitudinal neural decoding, focusing on the amount of data required for pretraining and adaptation to the target day.

\subsection{Pretraining Data Requirements}
\label{app:pretraining_scale}

We investigate how the amount of historical pretraining data affects decoding performance after adaptation to the target day. {\name} is pretrained on the most recent $N \in \{5,10,20,39\}$
IBL sessions and evaluated on the same 14 held-out future sessions. The pretrained backbone is frozen, and linear probes are fitted using 20\% of each target day's data for labeled calibration. As shown in Figure~\ref{fig:pretraining_scale},
increasing the amount of pretraining data from 5 to 39 days improves movement decoding $R^2$ from
0.32 to 0.35 (+9.2\%) and choice balanced accuracy from 0.70 to 0.78 (+10.5\%).

\begin{figure}[htbp]
    \centering
    \includegraphics[width=\linewidth]
    {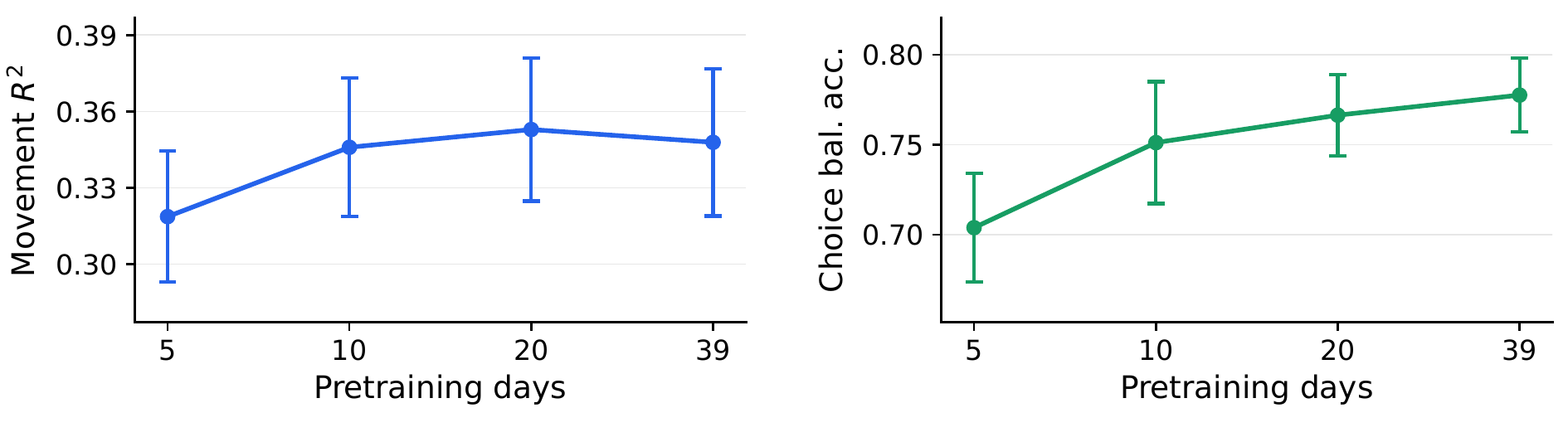}
    \caption{
    Effect of pretraining data size on decoding performance after adaptation to the held-out day using a frozen backbone and linear probes. (a) Movement-related behavior decoding measured by $R^2$. (b) Choice decoding measured by balanced accuracy. Results are averaged across 14 held-out days, with error bars showing $\pm$ 1 standard error across sessions.}
    \label{fig:pretraining_scale}
\end{figure}

\subsection{Calibration and Adaptation Efficiency}
\label{app:calibration_efficiency}

We further investigate how the amount of labeled calibration data affects decoding performance after adaptation to the target day. Following the chronological IBL
split \ref{app:data}, we use {\name}
pretrained on 39 days and evaluate on the same 14 held-out future days. We vary the number of labeled
trials $n \in \{4,8,16,32,\mathrm{all}\}$, using the first
$n$ trials from each day's 20\% split for adaptation, while keeping the
10\% validation and 70\% test splits unchanged. We compare gradient-free (GF) adaptation (solid lines) with gradient-based (GB) adaptation (dashed lines) in Figure~\ref{fig:labeled_adaptation_efficiency}.

As shown in Figure~\ref{fig:labeled_adaptation_efficiency}, decoding performance improves with increasing amounts of labeled calibration data. With the full 20\% adaptation split, GF achieves a movement-related behavior $R^2$ of 0.32 and a choice balanced accuracy of 0.68, compared with 0.39 and 0.83 for GB, respectively. These results indicate that gradient-based adaptation provides additional gains over gradient-free approaches, particularly as more labeled trials become available.

\begin{figure}[htbp]
    \centering
    \includegraphics[width=\linewidth]
    {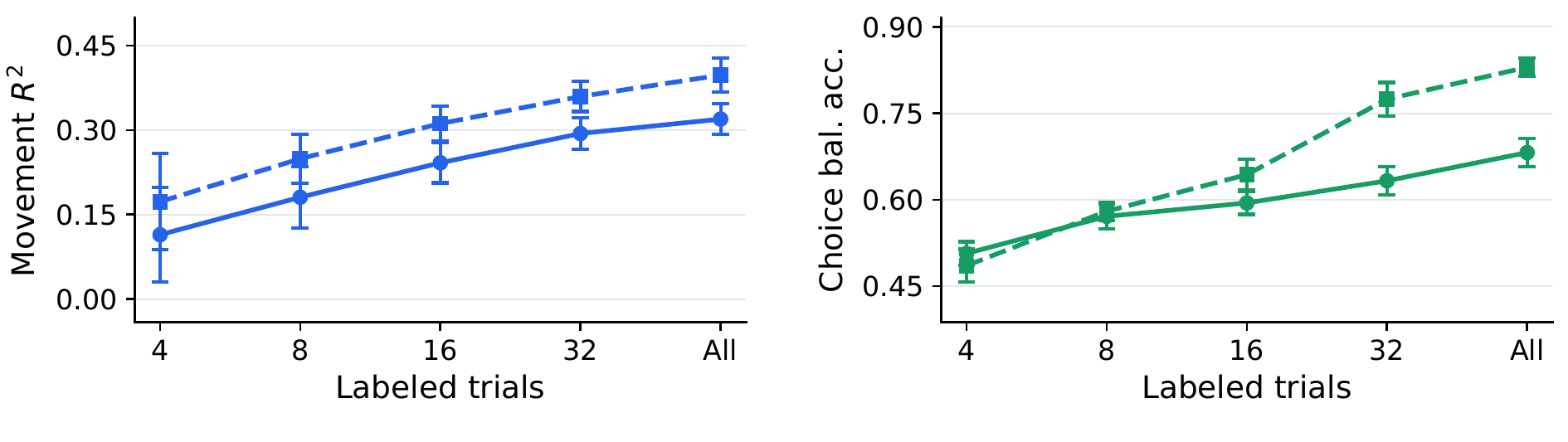}
    \caption{
    Effect of labeled calibration data size on decoding performance with gradient-free and gradient-based adaptation. (a) Movement-related  decoding measured by $R^2$. (b) Choice decoding measured by balanced accuracy. Solid lines denote gradient-free (GF) adaptation, and dashed lines denote gradient-based (GB) adaptation. Results are averaged across 14 held-out days, with error bars indicating $\pm$ 1 standard error across sessions.
    }
    \label{fig:labeled_adaptation_efficiency}
\end{figure}

\par
\endgroup

%% file: appendix/2-stage_training.tex
\section{Two-Stage Model Training}\label{app:2_stage_training}

For \textit{generalization across time} (Figure~\ref{fig:figure3}), all baselines are trained in 2 stages: (1) In \textbf{stage~1}, all models are trained on the same pretraining data corpus: SSL models are not trained on any labels, whereas supervised baselines are pretrained with labels; (2) \textbf{stage~2} adapts the pretrained model to a target day or participant with supervision via finetuning, which is the same procedure for all baselines. Table~\ref{tab:stage_overview} summarizes the detailed comparisons.

\begin{table*}[h!]
\centering
\caption{Two-stage training setup used for both the IBL and human chronic datasets, across all baselines, for evaluating generalization across time.}
\label{tab:stage_overview}
\small
\renewcommand{\arraystretch}{1.3}
\begin{tabular}{@{}p{0.10\textwidth}p{0.42\textwidth}p{0.42\textwidth}@{}}
\toprule
 & \textbf{IBL} & \textbf{Human} \\
\midrule
\textbf{Stage 1} &
39 training days, with labels used only for supervised baselines &
Pooled T12 and T15 training data, with labels used only for supervised baselines \\
\addlinespace
\textbf{Stage 2} &
Backbone frozen; only update the read-in layers required to map new neural populations to the latent space, and the task-specific decoder heads & Full end-to-end finetuning (updating the backbone, read-in, and decoder heads) \\
\bottomrule
\end{tabular}
\end{table*}

%% file: appendix/learning.tex
\begingroup

\section{Learning-related analyses}\label{app:learning}

This appendix describes the analyses in Figure~\ref{fig:figure4}, using chronic recordings from one mouse spanning initial training through expert performance \citep{melin2026behavioral}. The model was pretrained on 39 days of neural recordings without behavioral labels. Held-out evaluation sessions from the remaining days were excluded from these analyses.

\paragraph{Learning state discovery.} We compared task-related representations using neural activity and {\name} latents across sessions (Figure~\ref{fig:figure4}A). We computed squared Euclidean distances between all pairs of condition-averaged population vectors, corrected for within-session trial-sampling noise and normalized by dimensionality. These distances across conditions and time bins defined a task-geometry vector for each session. Cross-day similarity was calculated as the Pearson correlation between these geometry vectors. The resulting heatmaps revealed three distinct clusters of sessions. We then examined behavior and task variables, including stimulus direction and contrast, choice, and feedback following correct choices, to characterize these clusters as three learning stages: \textit{debiasing}, \textit{novice}, and \textit{expert} (Figure~\ref{fig:figure4} B--D).

\paragraph{Behavioral strategy contributions.}
Adapting the approach of \citet{sun2025learning}, we quantified how strongly each classified learning stage contributed to the animal's behavioral profile. We summarized each session by its wheel-velocity profiles across stimulus sides and contrasts, and fitted these jointly using three stage-specific templates. Each template was averaged over representative sessions from that learning stage, excluding the session being evaluated. To quantify a template's unique contribution, we removed it, refitted the remaining templates, and calculated the proportional increase in residual error relative to the reduced model. We normalized these coefficients of partial determination (CPDs) by their sum to track the relative contributions of the three stages across learning, and estimated transition widths from sigmoid fits to these trajectories.

\paragraph{Choice-related neural structure.}
We extracted neural activity and {\name} latents from the same trials and temporal bins within each session for comparison. To isolate choice-related activity, we fitted a regularized linear encoding model for each session using stimulus, choice and feedback, together with wheel speed, licking and whisker motion. Regularization was selected by cross-validation, and predictions were evaluated on held-out trials. Adapting the population-vector analysis of \citet{sun2025learning}, we averaged the choice-predicted component separately across left- and right-choice trials, forming a population vector across units or latent dimensions at each time bin. We computed Pearson correlations between these vectors for every pair of bins along a concatenated stimulus- and movement-aligned time axis (Figure~\ref{fig:figure4}F). Heatmap rows and columns correspond to left- and right-choice trial times, respectively. Each heatmap matrix was averaged within each learning stage. Lower correlations indicate more distinct choice-related population patterns.

\par

\endgroup

%% file: appendix/model.tex
\section{Model and Hyperparameter Details}
\label{app:model_details}

\subsection{Model Capacity Control via Shared Backbone}

In our experiments, we controlled backbone capacity across all baselines. Transformer-based methods (e.g., {\name}, causal NDT, SPINT, POYO) used a common backbone configuration, matched in depth, width, and number of attention heads (Table~\ref{tab:shared_backbone_and_sizes}). Therefore, differences in downstream performance among these methods reflect their training objectives and model-specific design rather than backbone capacity. For non-transformer baselines (e.g., POSSM, CEBRA), for which a shared backbone architecture was not applicable, we matched the total number of backbone parameters to that of the transformer backbone (Table~\ref{tab:shared_backbone_and_sizes}). Model backbone hyperparameters were selected independently for the IBL and human datasets.

\begin{table*}[h!]
\centering
\caption{Shared Transformer backbone architecture and backbone size for the IBL and human chronic datasets. Backbone size includes the causal self-attention blocks but excludes model-specific encoders that project spikes to latents before the backbone, decoder heads, and other model-specific components.}
\small
\setlength{\tabcolsep}{7pt}
\renewcommand{\arraystretch}{1.15}
\begin{tabular}{|l|cc||l|cc|}
\hline
\multicolumn{3}{|c||}{\textbf{Shared Transformer Backbone}}
&
\multicolumn{3}{c|}{\textbf{Backbone Parameters}} \\
\hline
\textbf{Hyperparameter} & \textbf{IBL} & \textbf{Human}
&
\textbf{Model} & \textbf{IBL} & \textbf{Human} \\
\hline
Layers                    & 6      & 6
& {\name}                 & 3.58M  & 7.29M \\
Embedding dimension       & 192    & 384
& Causal NDT              & 3.28M  & 7.29M \\
Attention heads           & 8      & 6
& POSSM                   & 3.57M  & 7.27M \\
FFN dim                   & 1024   & 768
& SPINT                   & 3.55M  & ---   \\
Activation                & GELU   & GELU
& CEBRA                   & 0.31M  & 0.44M \\
Normalization             & Pre-LN & Pre-LN
& POYO+                   & 3.55M  & ---   \\
Dropout                   & 0.2    & 0.2
&                         &        &       \\
\hline
\end{tabular}
\label{tab:shared_backbone_and_sizes}
\end{table*}

\subsection{Model-Specific Details}
\label{app:model_arch}

This subsection describes the components that distinguish each model from the shared backbone (Table~\ref{tab:shared_backbone_and_sizes}): how neural activity is encoded, the pretraining objective, and the latent representation used for downstream decoder heads.

\paragraph{Linear baselines.}
We use Ridge regression to decode time-bin-level behavior and logistic regression for trial-level choice, using \textit{scikit-learn} packages, and choose hyperparameters, e.g., the $L_2$ penalty, based on cross-validation. The same approach is used both for decoding with the raw spike baselines and for linear probing with the learned neural latents (Figure~\ref{fig:figure2}). For time-bin-level behavior, the neural activity and behavior are time-bin aligned. Neural activity inputs are standardized on the training split, and the Ridge regression is fit independently per behavioral variable. For the trial-level choice variable, the neural latents are averaged over all time bins and pooled into a single vector for decoding. For the raw spike baseline, the decoder takes in neural activity across all timesteps and concatenates them as input to the logistic regression.

\begin{table}[ht!]\centering
\caption{Model-specific hyperparameters used for {\name}.}
\begin{tabular}{|l|c|c|}
\hline
\textbf{Hyperparameter} & \textbf{IBL} & \textbf{Human} \\ \hline
Input representation & Binned spike counts & 5-bin time patches \\
Input dimension & $N$ neurons & 1280 (T12); 2560 (T15) \\
Latent dimension & 192 & 384 \\
Attention heads & 8 (predictor) \& 1 (encoder) & 6 \\
FFN dimension & 1024 (predictor) \& 256 (encoder) & 768 \\
Dropout & 0.1 & 0.1 \\
Day adaptation & Neuron identity embeddings & Per-Day FiLM \\
SIGReg loss weight & 0.6 & 1.0 \\
Context window size & 59 preceding bins & All preceding patches \\
Backbone \# of parameters & 3.58M & 7.29M \\ \hline
\end{tabular}
\label{tab:neurolens_architecture}
\end{table}

\paragraph{NeuroLens.} 
NeuroLens is our self-supervised framework for learning transferable
neural representations across recording sessions. For calibration-based adaptation, we sample 64 calibration time windows to infer neuron embeddings on the IBL dataset. On human chronic datasets, the number of calibration time windows used vary across days as recording sessions from some days are fewer than 64. 
For trial-level decoding (choice and sentence embedding), we average
the neural latents from the predictor outputs
$\widehat{\mathbf{z}}_{1:T}^{(d)}$ across the $T$ time bins. On IBL,
a $\mathrm{LayerNorm}\rightarrow\mathrm{Dropout}\rightarrow
\mathrm{Linear}(192,2)$ head predicts choice with a binary
cross-entropy loss; on the human chronic datasets, a
$\mathrm{LayerNorm}\rightarrow\mathrm{Linear}(384,384)$ head
predicts the sentence embedding with a cosine similarity loss. For
time-bin-level variables, predictions are made from $\mathbf{h}_t$
at each time bin: on IBL, a
$\mathrm{LayerNorm}\rightarrow\mathrm{Linear}(192,192)
\rightarrow\mathrm{GELU}\rightarrow\mathrm{Dropout}
\rightarrow\mathrm{Linear}(192,6)$ head regresses the
movement-related behaviors with a mean-squared-error loss; on the
human chronic datasets, a
$\mathrm{LayerNorm}\rightarrow\mathrm{Linear}(384,41)$ head produces
phoneme logits, trained with a CTC loss.

\begin{table}[ht!]\centering
\caption{Model-specific hyperparameters used for Causal NDT.}
\begin{tabular}{|l|c|c|}
\hline
\textbf{Hyperparameter} & \textbf{IBL} & \textbf{Human} \\ \hline
Input representation & Binned spike counts & 5-bin time patches \\
Latent dimension & 192 & 384 \\
\# of attention heads & 8 & 6 \\
FFN dimension & 1024 & 768 \\
Dropout & 0.2 & 0.2 \\
Context window size & 59 preceding bins & All preceding patches \\
Backbone \# of parameters & 3.28M & 7.29M \\ \hline
\end{tabular}
\label{tab:ndt_architecture}
\end{table}

\paragraph{Causal NDT.} 
Causal NDT is a self-supervised causal Transformer baseline built on \citep{ye2021representation}. A 2-layer MLP read-in maps neural activity at each time step to the latent space, a  causal transformer backbone (Table~\ref{tab:shared_backbone_and_sizes}) processes the sequence, and a linear output layer maps each hidden state back to observation space to predict the next step's activity. For IBL, inputs are raw spike counts. For multi-day pretraining, each session (day) has its own read-in and output layer during stage-1 training, since each day has a distinct number of neurons. The output layer predicts per-neuron log firing rates, trained to predict the next bin's spike counts under a Poisson negative log-likelihood, with no behavioral labels used. In stage 2, only a newly initialized per-day read-in and the decoding heads are updated, with the rest of the model frozen. For decoding, the decoder heads and losses are the same as other baselines (binary cross-entropy for choice, MSE for time-bin-level behavior). For the human chronic dataset, 5 consecutive 20 ms time bins are concatenated along the channel dimension to form 1 input time patch. The spike data are projected with a 2-layer MLP followed by a learned temporal positional embedding, and we employ separate participant-specific output layers to reconstruct their own spike data, while keeping the Transformer backbone shared. In Stage 1, the training objective is next-step prediction of the following time patch's neural features using an MSE loss; no behavioral labels are required for this stage, and the spike reconstruction output layer is discarded after Stage 1. In Stage 2, the model is fine-tuned end-to-end on a single participant with the task-specific decoding losses, as in Appendix~\ref{app:phoneme_semantic_decoding}.

\begin{table}[ht!]
\caption{Model-specific hyperparameters used for POSSM. }
  \centering
  \begin{tabular}{|l|c|c|}
  \hline
  \textbf{Hyperparameter} & \textbf{IBL} & \textbf{Human} \\ \hline
  Input representation & Spike events & 5-bin time patch \\
  Latent dimension & 256 & 192 \\
  \# of latents per time step & 1 & 1 \\
  \# of cross-attention heads & 1 & 1 \\
  Cross-attention head dimension & 32 & 64 \\
  GRU depth & 2 & 8 \\
  GRU hidden dimension & 588 & 395 \\
  GRU backbone \# of parameters & 3.57M & 7.27M \\
  Is current input time step included? & No & No \\
  Context window size & 59 preceding states & All preceding states \\ \hline
  \end{tabular}
  \label{tab:possm_architecture}
\end{table}

\paragraph{POSSM.}
We implement POSSM \citep{ryoo2026generalizable} as a supervised
multi-session baseline with an event-based neural read-in, a
unidirectional GRU temporal backbone, and a causal cross-attention
decoder.
Unlike the shared IBL input representation, which consists of vectors
of binned spike counts, POSSM retains individual spike events within
each 16.7 ms bin.
Each spike is represented by a learned unit embedding, and one learned
latent token cross-attends to the spike-event tokens within each bin.
The resulting bin representation is processed by the GRU, after which
an output query cross-attends to preceding GRU hidden states to produce
the downstream latent.

During multi-day IBL pretraining, each recording day has its own unit and day embeddings, while the input cross-attention, GRU, output cross-attention, and supervised task heads are shared across recording days. The stage-1 training objective is the equally weighted sum of time-bin-level (movement) behavior MSE and choice binary cross entropy losses. Both the time-bin-level movement-related behavior output and the choice decoder use linear projections of the 192-dimensional decoder representation (Table~\ref{tab:possm_architecture}).

To align POSSM with the strictly causal setting adopted by {\name}, each recurrent chunk contains exactly one 16.7 ms time bin, and the decoder excludes
the hidden state corresponding to the current bin.
When producing the latent for time bin $t$, the decoder explicitly attends to time bins up to time bin $t - 1$.
The unidirectional GRU may carry information from earlier states into this explicit context, but neither the GRU nor the decoder receives neural activity from the current or future bins to prevent data leakage. For an unseen IBL day, the recurrent and cross-attention modules remain frozen during target day adaptation, while the new day-specific unit embeddings and
the fresh decoder head are updated during few-shot finetuning.

For BCI, 5 consecutive 20 ms feature vectors are concatenated along the channel dimension to produce one 100 ms input step. T12 and T15 use separate participant-specific 2-layer MLP read-in layers, followed by a shared POSSM input cross-attention, GRU, causal decoder and decoder head.
The decoder head attends to all preceding recurrent states while excluding the current 100 ms input step.
The phoneme decoding model applies
$\mathrm{LayerNorm}\rightarrow\mathrm{Linear}(256, 41)$
to every decoded step.
The sentence embedding decoding model averages the decoded representations over all time steps and applies
$\mathrm{LayerNorm}\rightarrow\mathrm{Linear}(256, 384)$.
We separately train the model to decode phonemes and sentence embeddings because joint optimization caused substantial interference between the two tasks.

We perform a targeted architectural search rather than a large random hyperparameter sweep. For IBL, we compare explicit decoder contexts of 9 and 59 states and
multiple GRU depth, including width near the shared backbone budget. The reported model uses a four-layer GRU with hidden size 588. For T12 and T15, we compare multiple GRU depths while matching the full
7.29-million-parameter transformer backbone budget.
The reported model-capacity-aligned model uses 8 GRU layers with hidden size 395. The remaining architectural settings are summarized in Table~\ref{tab:possm_architecture}.

\begin{table}[h!]
\caption{Model-specific hyperparameters used for SPINT, which is only applied to IBL data.}
\centering
\begin{tabular}{|l|c|}
\hline
\textbf{Hyperparameter} & \textbf{IBL} \\ \hline
Input representation & Binned spike counts \\
Latent dimension & 256 \\
Cross-attention depth & 2 \\
\# of attention heads \& head dimension & 4 \& 64 \\
FFN dimension & 2944 \\
Activation & ReLU \\
Normalization & Pre-LN \\
Attention \& FFN \& residual dropout & 0.1 \\
Dynamic neuron dropout & $p\sim\mathrm{Uniform}(0.1,0.5)$ \\
\# of latent queries & 6 \\
\# of calibration trials per day & 64 \\
Is current input time step included? & No \\
Context window size & 59 preceding bins \\
Backbone \# of parameters  & 3.55M \\ \hline
\end{tabular}
\label{tab:spint_architecture}
\end{table}

\paragraph{SPINT.} We implement SPINT (\textit{Spatial Permutation-Invariant Neural
Transformer})~\citep{le2026spint} as a supervised multi-session decoding baseline on the IBL chronic dataset, with a calibration-based neural read-in module and spatial cross-attention over recorded units. For each neuron, the model takes its spike count sequence across consecutive time bins and represents that sequence as a single token. A shared identity network encodes 64 unlabeled calibration windows from a session's training span, averages the identity network outputs across windows, and maps them to a per-neuron identity vector. This neuron identity vector is added to each neuron's projected spike count token before the shared MLP read-in. Learned latent queries, one per behavior output, cross-attend to the resulting neural tokens to produce fixed-dimensional query latents. This read-in module supports variable-sized populations without a learnable neuron-identity lookup table. Dynamic neuron dropout is also employed to improve robustness to changes in the recorded population.

During multi-day pretraining, all model parameters are shared across days, while a calibration set is sampled from each day. For IBL, the stage-1 training objective is an equally weighted sum of mean-squared error for time-bin-level, movement-related behavior and binary cross-entropy for choice. The 5 time-bin-level behaviors comprise 6 scalar outputs because left-paw motion has 2 dimensions. A shared linear decoder head maps each query latent to a behavior sequence. For choice decoding, query latents are averaged across time steps within each trial and passed through a 2-layer MLP. Choice supervision is applied only to trials containing a stimulus onset and a valid choice label; all other time windows contribute only to time-bin-level behavior decoding. To align SPINT with the strictly causal decoding setup, for a window ending at time bin $t$, the prediction uses neural activity only up to time $t-1$, together with a fixed calibration set.

For an unseen day, neuron embeddings are inferred from calibration sets sampled from its adaptation split. The cross-attention backbone and learned queries remain frozen, whereas the identity encoder, shared spike read-in, and task-specific decoder heads are adapted. For time-bin-level behavior and choice decoding, models are initialized from the same pretrained checkpoint but fine-tuned separately for their respective targets. For trial-level choice decoding, query latents are averaged across all time steps within a trial to obtain a per-trial vector. The model hyperparameters are summarized in Table~\ref{tab:spint_architecture}. SPINT is evaluated only on the IBL dataset.

\begin{table}[h!]
\caption{Model-specific settings used for CEBRA. Parameter counts refer to the encoder, which serves as the backbone for this model.}
\centering
\begin{tabular}{|l|c|c|}
\hline
\textbf{Hyperparameter} & \textbf{IBL} & \textbf{Human} \\ \hline
Input representation & Binned spike counts & 5-bin time patch \\
Encoder & \texttt{offset10-model} & MLP read-in + 3 conv blocks \\
Embedding dimension & 192 & 192 \\
Conv block latent dimension & 128 & 192 \\
Receptive field & 10 time bins & 10 time patches \\
Positional embedding & None & None \\
Distance \& temperature & Cosine \& auto (floor at 0.1) & Cosine \& fixed $\tau=1$ \\
Pretraining batch size & 512 & 64 \\
Finetuning batch size & 16 & 64 \\
Finetuning learning rate & $1 \times 10^{-3}$ & $5 \times 10^{-4}$ \\
Is current input time step included? & No & No \\
Backbone \# of parameters & 0.31M & 0.44M \\ \hline
\end{tabular}
\label{tab:cebra_architecture}
\end{table}

\paragraph{CEBRA.} CEBRA~\citep{schneider2023learnable} is a self-supervised contrastive learning framework that learns embeddings preserving temporal proximity, trained via an InfoNCE objective, wherein each anchor is paired with a positive sample from the same trial at a small temporal offset and contrasted against negatives sampled from the remaining anchors in the batch using cosine similarity, without any behavioral labels.

CEBRA is not transformer-based, so it does not use the shared backbone of Table~\ref{tab:shared_backbone_and_sizes}; its encoder is a convolutional network with a 10-step receptive field and no positional embedding. We adopt the architecture provided by the authors~\citep{schneider2023learnable} rather than widening it to match the transformer parameter count. On IBL data, we use the provided model config \texttt{offset10-model} (128 hidden units, 192 latent dimensions). On T12 and T15, we employ the same 2-layer MLP read-in as for the other baselines, followed by 3 causal convolutions (kernel 4, receptive field 10 patched steps, width 192), so that it shares the participant-specific read-in and the decoder heads with the other baselines. Because the encoder has no positional embedding and no cross-time mixing beyond its receptive field, its per-time-bin embedding is window-invariant, which the other models' positional embeddings do not permit.

The provided encoder centers its receptive field, so the embedding labelling time bin $t$ would read time bins $[t-5, t+5)$ (4 bins into the future). We shift the receptive field to end 1 bin early, so that we use neural activity only up to $t - 1$, to prevent data leakage and match the strictly causal setup under which the other models are scored. Downstream, the per-time-bin embedding is the latent: time-bin-level behaviors are decoded per time bin, and the trial-level target is decoded from the embedding averaged over the time bins within a trial, then classified with a linear decoder. On an unseen day, the pretrained backbone is frozen, and that day's own read-in encoder is trained together with a new read-in layer to accommodate the new data distribution, which aligns with CEBRA's own few-shot update setting ~\citep{schneider2023learnable}.

For few-shot finetuning adaptation, we searched the learning rate ($1 \times 10^{-3}$, $3 \times 10^{-4}$, $1 \times 10^{-4}$) and batch size (16, 64) on 3 held-out sessions spanning the range of population sizes; during pretraining, we searched the embedding dimension (32, 192). In both cases, we selected the best checkpoint based on validation decoding performance. The other settings are summarized in Table~\ref{tab:cebra_architecture}.

\begin{table}[h!]
\caption{Model-specific settings used for POYO.}
\centering
\begin{tabular}{|l|c|}
\hline
\textbf{Hyperparameter} & \textbf{IBL} \\ \hline
Input representation & Spike events \\
Latent dimension & 192 \\
\# of latents per time step & 2 \\
Latent grid step & 1 bin (16.7\,ms) \\
\# of cross-attention heads \& head dimension & 2 \& 32 \\
Time positional embedding & RoPE \\
Decoder head & 2-layer MLP (width 192, dropout 0.2) \\
Unit dropout & 0.1 \\
Finetuning learning rate \& weight decay & $3\times10^{-4}$ \& 0.05 \\
Finetuning batch size & 64 \\
Is current input time step included? & No \\
Context window size & 59 preceding bins \\
Backbone \# of parameters & 3.55M \\ \hline
\end{tabular}
\label{tab:poyo_architecture}
\end{table}

\paragraph{POYO.} POYO~\citep{azabou2023unified} is a supervised decoder trained to decode behavioral targets. Unlike models based on binned spike counts, it tokenizes individual spikes, each with a per-neuron embedding and a rotary time embedding (RoPE); since timestamps enter the tokenizer for spikes rather than via learnable time embeddings, the backbone has no positional embedding, so its Table~\ref{tab:shared_backbone_and_sizes} count reflects only the self-attention blocks. A cross-attention read-in compresses spikes within each 16.7 ms step into latent tokens, the shared causal Transformer backbone processes them, and a cross-attention read-out queries the result at the output times before per-target MLP decoder heads. The objective equally weights behavior MSE and choice binary cross entropy.

Attention is masked on three levels so that no model output depends on neural activity after the time bin it is queried at (no temporal leakage): (1) the cross-attention read-in latent attends only to spikes preceding the end of its own time step, (2) latents attend only to earlier latents, (3) and the cross-attention read-out query attends only to latents whose time step has ended. For an unseen day, the read-in, backbone, and read-out are frozen, and only that session's new unit and day embedding rows are updated together with a fresh decoder head, following POYO's own few-shot recipe~\citep{azabou2023unified}, as also done for POSSM~\citep{ryoo2026generalizable}. For hyperparameter tuning, we searched the finetuning learning rate, batch size, window stride, unit dropout ratio, and different initializations for new rows in the neuron embedding lookup table, selecting the best checkpoint based on validation decoding performance. The other settings are summarized in Table~\ref{tab:poyo_architecture}. POYO is evaluated on IBL data only. 

%% file: appendix/training.tex
\section{Training Details}
\label{app:training_details}
All baselines are trained with the AdamW optimizer \citep{loshchilov2017decoupled} and employ the same training setup and budgets. The checkpoint selection approach, based on validation metrics, is also the same across all baselines to ensure fair comparisons. We describe training details for each baseline below.

\paragraph{\name.} We train both the single-day and multi-day {\name} using a learning rate of $1\times10^{-4}$ and weight decay 0.05. For IBL, the 39-day model is trained on a single Nvidia A40 GPU in 21.1 hours for a total of 1000 epochs. Single-day models can be trained on a single Nvidia A40 or A100 GPU in 1.0 $\sim$ 2.7 hours, also for 1000 epochs. Stage-2 fine-tuning (behavior and choice, 300 epochs each) requires 1.1 $\sim$ 4.0 hours per held-out session on one A40 or A100 GPU. For T12 and T15 datasets, {\name} is trained under all experimental settings on a single Nvidia A100 GPU (40\,GB) for a total of 1000 epochs: approximately 20 hours for the high- vs.\ low-level decoding setting and 16 hours for the generalization-across-time setting. Fine-tuning can be performed on a single A100 or A40 GPU for 500 epochs, requiring approximately 2 $\sim$ 3 hours per run for T12 and 6 $\sim$ 9 hours for T15. For the ablation experiments, we use one A100 or A40 GPU and complete each run in 2 $\sim$ 9 hours.



\paragraph{Causal NDT.} We train both the single-day and multi-day Causal NDT using the same optimizer settings as {\name}. For IBL, the 39-day model is trained on a single Nvidia A100 GPU in 8.7 hours for a total of 1000 epochs. Single-day models can be trained on a single Nvidia A40 or A100 GPU in 0.5 $\sim$ 1.4 hours, also for 1000 epochs. Stage-2 fine-tuning (behavior and choice, 300 epochs each) requires 1.2 $\sim$ 3.8 hours per held-out session on one A40 or A100 GPU. For T12 and T15 datasets, causal NDT is trained under all experimental settings on a single Nvidia A100 GPU (40 GB) for a total of 1000 epochs: approximately 16 hours for the high- vs.\ low-level decoding setting and 14 hours for the generalization-across-time setting. Fine-tuning can be performed on a single A100 or A40 GPU for 500 epochs, requiring approximately 2.4 $\sim$ 3 hours per run for T12 and 5.5 $\sim$ 8 hours for T15; fine-tuning repeats over random seeds fall in the same range (2.4 $\sim$ 2.6 hours for T12, 5.5 $\sim$ 6.2 hours for T15).

\paragraph{POSSM.} We train both the single-day and multi-day POSSM using the AdamW optimizer. On IBL, the 39-day model is pretrained on a 96GB H100 GPU in $\sim$9.8 hours for a total of 1000 epochs with early stopping, using batch size 64, peak learning rate 0.001, weight decay 0.05, unit dropout 0.15, gradient clipping at 1, and 15\% warmup for the learning rate scheduler. Stage-2 finetuning is run on a single GPU in $\sim$1 hour using two sub-stages: (1) unit identification (learning rate $3\times10^{-4}$, at most 100 epochs) followed by (2) read-in  and readout fine-tuning (learning rate $1\times10^{-5}$, at most 400 epochs), both with batch size 16 and weight decay 0.1. On T12 and T15 data, POSSM is trained on a single H100 GPU in $\sim$2.5 hours for a total of 500 epochs, using AdamW with batch size 32, peak learning rate $5\times10^{-4}$, minimum learning rate $1\times10^{-5}$, weight decay $10^{-3}$, cosine decay with 2\% warmup, and gradient clipping at 10.

\paragraph{SPINT.} On IBL data, SPINT is pretrained for 1000 epochs on the 39 pooled days. Training uses AdamW optimizer with a OneCycle learning-rate scheduler, with each batch drawn from a single day. The Stage-1 checkpoint maximizes the mean per-behavior validation $R^2$, computed from predictions pooled across the pretraining days' validation set. Choice cross-entropy contributes to pretraining but does not enter this model selection criterion. For finetuning, we employ training batches of 16 trials. The reported choice decoding runs use AdamW optimizer with peak learning rate $1\times10^{-4}$, weight decay 0.05, a OneCycle schedule with 15\% warmup, and gradient clipping at 1.0, for a fixed budget of 500 epochs. Training runs for the full epoch budget without early stopping.

\paragraph{CEBRA.} We train CEBRA using its own solver for Stage~1 and AdamW for Stage~2. On IBL, Stage~1 is trained on one A40 or A100 GPU in $\sim$1 hour for a total of 12{,}000 iterations, on the 39 pooled days, with batch size 512 and learning rate $3\times10^{-4}$. Stage~2 is trained on a single GPU in 2$\sim$14 minutes per test day per behavior target, attaching a decoder head to the frozen pretrained encoder and training it together with the held-out day's own read-in encoder, using learning rate $1 \times 10^{-3}$, weight decay 0.05, batch size 16, with early stopping at up to 1000 epochs. Time-bin-level behavior and choice decoding are independent runs from the same Stage-1 checkpoint, selected on validation mean $R^2$ and validation balanced accuracy respectively. On T12 ad T15 datasets, CEBRA follows the shared two-stage training protocol in Table~\ref{tab:stage_overview}. Stage~1 is trained on one A40 or A100 GPU in $\sim$7.3 hours for a total of 1000 epochs, pooling T12 and T15 with learning rate $5\times10^{-4}$, weight decay $1 \times 10^{-5}$, cosine schedule without warmup, batch size 64, using a time-contrastive loss with a fixed temperature of 1. Stage~2 is trained on a single GPU for a total of 1000 epochs, once per participant and behavioral variable, fine-tuning end to end and taking $\sim$3.7 hours per run for T12 and 7 hours for T15.

\paragraph{POYO.} We train POYO using the AdamW optimizer. On IBL data, Stage~1 is trained on one A40 or A100 GPU in 2.8 $\sim$ 5.2 hours across runs, for 1000 epochs with early stopping on the 39 pooled days, using learning rate $3\times10^{-4}$, weight decay 0.05, batch size 64, gradient clipping at max-norm 1.0, and a cosine schedule with 5\% warmup. The training objective equally weights time-bin-level behavior MSE and choice binary cross-entropy. Stage~2 is trained on a single GPU in $\sim$25 seconds to 5 minutes per test day per behavior target; finetuning runs are short since they early-stop well within the epoch budget. We freeze the read-in, backbone, and read-out, training only the held-out day's new unit and session embedding rows with a new decoder head, using learning rate $3\times10^{-4}$, weight decay 0.05, and batch size 16, for at most 500 epochs. Time-bin-level and choice decoding are independent runs, selected by validation mean $R^2$ and balanced accuracy, respectively.